\documentclass[nohyperref]{article}

\usepackage{microtype}
\usepackage{graphicx}
\usepackage{subfig}
\usepackage{booktabs} % for professional tables
\usepackage{pdfx}

\usepackage{hyperref}

\usepackage[accepted]{icml2022}

\usepackage{amsmath}
\usepackage{amssymb}
\usepackage{mathtools}
\usepackage{amsthm}

\usepackage[capitalize,noabbrev]{cleveref}

\usepackage{float} 
\usepackage{amsmath,amsthm,amssymb}
\usepackage{bm,upgreek}
\usepackage{hyperref}
\usepackage{tikz}
\usetikzlibrary{quotes}
\usetikzlibrary{bayesnet}
\usetikzlibrary{shapes,arrows}
\usepackage{booktabs}       % professional-quality tables
\usepackage[tableposition=top]{caption}

\usepackage{soul}
\usepackage{placeins}

\theoremstyle{plain}
\newtheorem{theorem}{Theorem}[section]
\newtheorem{proposition}[theorem]{Proposition}
\newtheorem{lemma}[theorem]{Lemma}

\theoremstyle{definition}

\theoremstyle{plain}
\newtheorem{remark}[theorem]{Remark}

\icmltitlerunning{Imitation Learning by Estimating Expertise of Demonstrators}
\newcommand{\States}{\mathcal{S}}

\newcommand{\Actions}{\mathcal{A}}
\newcommand{\reals}{\mathbb{R}}
\newcommand{\demons}{\mathcal{D}}

\newcommand{\ileed}{ILEED}
\newcommand{\loss}{\mathcal{L}}

\newcommand{\fembed}{f_{\phi}}

\newcommand{\bomega}{\omega}

\newcommand{\skill}{\sigma}

\newcommand{\numexperts}{m}
\newcommand{\dimembed}{d}
\begin{document}

\twocolumn[
\icmltitle{Imitation Learning by Estimating Expertise of Demonstrators}

% It is OKAY to include author information, even for blind
% submissions: the style file will automatically remove it for you
% unless you've provided the [accepted] option to the icml2022
% package.

% List of affiliations: The first argument should be a (short)
% identifier you will use later to specify author affiliations
% Academic affiliations should list Department, University, City, Region, Country
% Industry affiliations should list Company, City, Region, Country

% You can specify symbols, otherwise they are numbered in order.
% Ideally, you should not use this facility. Affiliations will be numbered
% in order of appearance and this is the preferred way.
\icmlsetsymbol{equal}{*}

\begin{icmlauthorlist}
\icmlauthor{Mark Beliaev}{equal,ucsb}
\icmlauthor{Andy Shih}{equal,stanford}
\icmlauthor{Stefano Ermon}{stanford}
\icmlauthor{Dorsa Sadigh}{stanford}
\icmlauthor{Ramtin Pedarsani}{ucsb}
\end{icmlauthorlist}

\icmlaffiliation{ucsb}{Department of Computer Science, University of California, Santa Barbara}
\icmlaffiliation{stanford}{Department of Computer Science, Stanford University}

\icmlcorrespondingauthor{Mark Beliaev}{mbeliaev@ucsb.edu}
\icmlcorrespondingauthor{Andy Shih}{andyshih@cs.stanford.edu}

% You may provide any keywords that you
% find helpful for describing your paper; these are used to populate
% the "keywords" metadata in the PDF but will not be shown in the document
\icmlkeywords{Machine Learning, ICML}

\vskip 0.3in
]

% this must go after the closing bracket ] following \twocolumn[ ...

% This command actually creates the footnote in the first column
% listing the affiliations and the copyright notice.
% The command takes one argument, which is text to display at the start of the footnote.
% The \icmlEqualContribution command is standard text for equal contribution.
% Remove it (just {}) if you do not need this facility.

%\printAffiliationsAndNotice{}  % leave blank if no need to mention equal contribution
\printAffiliationsAndNotice{\icmlEqualContribution} % otherwise use the standard text.

\begin{abstract}
Many existing imitation learning datasets are collected from multiple demonstrators, each with different expertise at different parts of the environment. Yet, standard imitation learning algorithms typically treat all demonstrators as homogeneous, regardless of their expertise, absorbing the weaknesses of any suboptimal demonstrators. In this work, we show that unsupervised learning over demonstrator expertise can lead to a consistent boost in the performance of imitation learning algorithms. We develop and optimize a joint model over a learned policy and expertise levels of the demonstrators. This enables our model to learn from the optimal behavior and filter out the suboptimal behavior of each demonstrator. Our model learns a single policy that can outperform even the best demonstrator, and can be used to estimate the expertise of any demonstrator at any state. We illustrate our findings on real-robotic continuous control tasks from Robomimic and discrete environments such as MiniGrid and chess, out-performing competing methods in \(21\) out of \(23\) settings, with an average of \(7\%\) and up to \(60\%\) improvement in terms of the final reward.
% Many existing imitation learning datasets are collected from multiple demonstrators, each with different expertise at different states of the environment. Yet, standard imitation learning algorithms typically treat the demonstration datasets as homogeneous, mimicking both the strengths and weaknesses of each demonstrator. In this work, we show that unsupervised learning over demonstrator expertise can lead to a consistent boost in the performance of imitation learning algorithms. We develop and optimize a joint model over a learned policy and expertise levels of the demonstrators. This enables our model to learn from the optimal behavior and filter out the suboptimal behavior of each demonstrator. Our model learns a single policy that can outperform even the best demonstrator, and can be used to estimate the expertise of any demonstrator at any state. We illustrate our findings on continuous control tasks and discrete environments such as minigrid and chess, out-performing competing methods in \(19\) out of \(21\) settings, with an average of \(7\%\) and up to \(60\%\) improvement in terms of the final reward.
\end{abstract}
\begin{figure*}[!t]
    \centering
    \includegraphics[width=\linewidth]{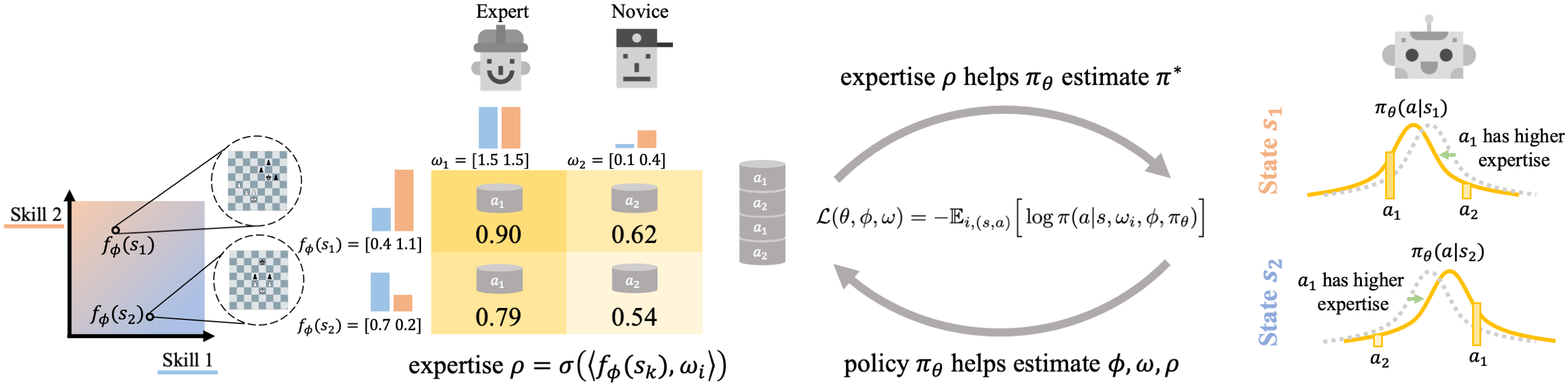}
    \caption{Left: State embeddings encode the skills associated with states of the environment. Middle: A demonstrator's expertise \(\rho\) at a state is a function of the state embedding and the demonstrator embedding.
    Right: Using the expertise levels, the model improves the learned policy \(\pi_\theta\), which in turn helps better estimate the state/demonstrator embeddings ($\phi$ and $\omega$) and the expertise level ($\rho$).}
    \label{fig:cover}
\end{figure*}
\section{Introduction}\label{sec:Introduction}
%1st paragraph: Impact of a problem and its challenges. Give very concrete real-world examples

Reinforcement learning provides a powerful and general framework for tasks such as autonomous vehicles, assistive robots, or conversation agents, by optimizing behavior with respect to user-specified reward functions. However, online interaction with the environment can be costly or even unsafe~\cite{sutton2018reinforcement,mihatsch2002risk,hans2008safe,garcia2015comprehensive}, and specifying reward functions can be difficult in practice~\cite{Hadfield-Menell17inverse}. 
%In recent years, we have witnessed many intelligent artificial agents such as autonomous vehicles, assistive robots, or conversation agents that are limited in their behavior as they lack the ability to act in a human-like manner. This substantially increases the need for mimicking human behavior. In these settings, one cannot simply rely on traditional Reinforcement Learning (RL) algorithms that \textit{reinforce} the learner's behavior using a reward signal, as this reward signal can be sparse or misspecified~\cite{Hadfield-Menell17inverse}, the environment itself can be costly to simulate~\cite{sutton2018reinforcement}, or receiving negative reward can be unsafe and unrealistic~\cite{mihatsch2002risk,hans2008safe,garcia2015comprehensive}.
% Instead, recent methods mitigate these issues by viewing the problem through the lens of offline learning, where the learner is only provided demonstrations of the desired behavior they are tasked to emulate~\cite{pinto2016supersizing,codevilla2018end}. Specifically, we focus on imitation learning (IL), where demonstrations consist of state-action pairs~\cite{pomerleau1991efficient,argall2009survey}.
Instead, one can mitigate these issues by viewing the problem through the lens of offline learning, where the learner either has access to demonstrations of the task along with the corresponding reward values as in offline reinforcement learning~\cite{levine2020offline}, or only has access to expert demonstrations without any reward information as in imitation learning~\cite{pomerleau1991efficient,argall2009survey}. In this work, we focus on the imitation learning setting---only assuming access to demonstrations.
%, and no other annotation.\par

%2nd paragraph: How the most relevant related work has addressed the problem, and how they are limiting. By the end of this paragraph, the reader should very clearly know what you want to solve and be concerned about it because they are convinced it is important.
The success of offline methods crucially depends on the availability of a large and diverse dataset~\cite{pinto2016supersizing, fu2020d4rl} and as such, there has been a flurry of work in collecting vast amounts of expert demonstrations in various domains~\cite{sharma2018multiple,zhang2018deep,mandlekar2019scaling,fu2020d4rl,robomimic2021}.
% , especially in the field of robotics, which can significantly benefit from imitating expert demonstrations, but is very much limited in terms of access to large expert demonstration datasets
A common finding from these works is the need for crowd-sourced data collection, both for scale and for diversity~\cite{sharma2018multiple}.
%These works share a common finding: diversity trumps quality, and it is better to collect large datasets from a diverse group of demonstrators~\cite{sharma2018multiple} instead of 
%A common finding from these works is that the path forward for learning imitating policies is to not restrict ourselves to limited expert datasets and instead collect large datasets from a diverse group of demonstrators~\cite{sharma2018multiple}. 
For example, Roboturk~\cite{mandlekar2019scaling} reports data-collection from \(54\) different humans, and Robomimic~\cite{robomimic2021} organizes its data into \(6\) different types of demonstrators with varying levels of expertise. Other works such as CARLA~\cite{dosovitskiy2017carla} use a mix of human and autonomous agents for collecting demonstrations.

These crowd-sourced data collection pipelines inevitably generate a diverse dataset of behavior from users with varying levels of expertise. 
%We have seen that many existing offline datasets consist of a heterogeneous mix of demonstrations from different demonstrators with varying expertise. 
\textit{Yet, current imitation learning algorithms typically treat these datasets as homogeneous.} 
By assuming all demonstrations are equally optimal, these algorithms may be blindly learning from the weaknesses of suboptimal demonstrators.
%Blindly imitating these large offline datasets can lead to undesirable behaviour considering the varying levels of optimality in the provided demonstrations. 
Instead, our work asks the question: can we make use of the knowledge that the trajectories come from different demonstrators with various levels of suboptimality? More specifically, can we use the information of which demonstrator provided which demonstration to improve learning?\par
% That is, we do \textit{not} know which demonstrators are more skillful, or even how many demonstrators there are at each skill level. However, we do know which trajectory was provided by which demonstrator -- a piece of information that is available in many existing datasets, and is also extremely easy to collect as part of crowd-sourcing pipelines.\par
%s3rd paragraph: Key Insight!! Here you want to tell the reader what is the key insight you had for solving the problem in a couple of sentences. The key insight is not “what” you have done. It is the “insight” part of it. What new ideas are used in this paper that pushes us beyond the already existing related work. If you can make the insight compelling and tangible. Feel free to explicitly say: “Our key insight is …” or put it in quotes or emphasize/italicize it, to draw attention to it.

% \emph{Our key insight is that an imitating policy should take into account the varying optimality of trajectories collected from a mixture of demonstrators}.
\emph{Our key insight is that imitation learning frameworks should account for the varying levels of suboptimality in large offline datasets by leveraging information about demonstrator identities.} 
%Although we are learning from trajectories without any labels or rewards, we can still leverage our knowledge of the demonstrator identities. 
For example, say we learn to play chess by imitating from a large dataset of games. We do not know the expertise levels of the players, but we do know which player played which games. Some players could be highly skilled Grandmasters, in which case we should treat their demonstrations seriously, but some could be novices, in which case we might ignore their demonstrations. 
Although we are not given their expertise levels, we can rely on information about which player played which games to estimate their expertise levels in an unsupervised manner. These estimated expertise levels can then be used to more effectively learn a chess-playing policy.
%We obtain an unsupervised estimate of the quality of moves so that we can identify which demonstrations are coming from the novice player and which are coming from the Grandmaster, enabling us to more effectively learn a chess-playing policy.\par

%4th paragraph: Talk about the approach/algorithm proposed in this paper. If you’re using an acronym for your algorithm, feel free to mention it here. Give an idea of the core components of your algorithm/approach, and end the paragraph with a bulleted list of explicit contributions of this work.
We propose \ileed, \emph{Imitation Learning by Estimating Expertise of Demonstrators}, to imitate the trajectories in the dataset, while simultaneously learning to account for the different demonstrators' suboptimalities without any prior knowledge of their expertise~(Fig.~\ref{fig:cover}). \ileed~optimizes a joint model over a learned policy and expertise levels, and recovers not just a single expertise value for each demonstrator, but a state-dependent expertise value that reveals which demonstrators are better at acting in specific states. We provide our implementation of \ileed~online~\cite{Beliaev_ILEED_2022}.
%The takeaway of our method is simple but broadly applicable -- many existing imitation learning datasets are created from a mix of demonstrators with varying expertise, by relying only on the demonstrators' identities we can boost the performance of standard imitation learning algorithms.

Our main contributions are as follows:
% \begin{itemize}
% \itemsep0em
% \item We develop an imitation learning algorithm that jointly optimizes for an imitating policy and the expertise levels of the demonstrators.  The learned policy outperforms policies trained without taking into account demonstrator identities, and sometimes even policies trained only on high-quality demonstrations.
% \item We use the joint model to estimate the (state-dependent) expertise of each demonstrator. In settings with ground-truth expertise, we show that our estimates align closely with the true expertise values.
% \item We experimentally show the success of our method compared to standard baselines on 1) simulated datasets for grid-word, 2) human datasets for continuous control, and 3) human datasets for chess endgames.
% \end{itemize}
%
\begin{itemize}
\itemsep0em
\item We develop an imitation learning algorithm that jointly optimizes for an imitating policy and the expertise levels of the demonstrators. The joint model can estimate \emph{state-dependent} expertise of demonstrators, identifying which demonstrator can perform well in which states.
\item We theoretically show that our model generalizes standard behavioral cloning, and that our algorithm can recover the optimal policy via maximum likelihood if the suboptimal demonstrations align with our model's generative process.
\item We experimentally show the success of our method compared to standard baselines on 1) simulated datasets for grid-world, 2) human datasets for continuous control, and 3) human datasets for chess endgames. We empirically demonstrate that our learned policy outperforms policies trained without taking into account demonstrator identities, and is comparable to policies trained only on high-quality demonstrations.
\end{itemize}

\section{Related Work}\label{sec: Related Work}
%To put our work in perspective, we go over several related fields within the larger scope of machine learning. 
First we discuss imitation learning, an instance of offline learning without the use of reward information. 
%Though our work falls into the general field of imitation learning, we distinguish ourselves by being 
Since our work is concerned with learning from suboptimal demonstrations, we then discuss connections to modeling expertise in more general supervised learning settings.

\noindent \textbf{Imitation Learning.}
There is a large body of work which approaches offline learning by relying on expert trajectories composed solely of state-action pairs, avoiding the need for labeling with a reward signal~\cite{argall2009survey, pomerleau1991efficient, ross2011reduction, finn2016guided, ho2016generative, ding2019goal}. The main challenge for IL approaches is the reliance on access to large amounts of expert demonstrations. In addition to this challenge, many real-world applications require the execution of multiple skills, where expert demonstrations are even more limited. Several lines of work aim to solve this: multi-task and meta imitation learning~\cite{babes2011apprenticeship,dimitrakakis2011bayesian,hausman2017multi,li2017infogail} as well as few shot learning~\cite{duan2017one,james2018task,singh2020scalable} tackle the data efficiency problem in IL by considering both transfer learning to unseen skills and learning diverse sets of skills from multiple expert policies. Although similar to our setting due to the dependence on multiple demonstrators, unlike our method these approaches often assume oracle demonstrations, and in some cases access to online fine-tuning.

Stepping away from traditional imitation learning work that assumes access to oracle demonstrations, we are specifically concerned with learning from crowd-sourced data, where suboptimal demonstrations are unavoidable. Several works have considered this setting \cite{brown2019extrapolating,brown2020better,chen2020learning, zhang2021confidence,cao2021learning}, analyzing the impact of suboptimal demonstrations, and developing novel methods which can relax the amount of supervision required. Unlike our method, all of these approaches require environment dynamics to train on the attained reward signal, and in some cases knowledge about rankings over the set of demonstrations. We emphasize that our method does not rely on knowledge of environment dynamics or expertise rankings, and we leverage only the demonstrator identity of each demonstration.

We discuss two recent works that also address the suboptimal setting without utilizing environment dynamics. The first uses behavioral cloning (BC) to learn an ensemble policy directly from noisy demonstrations~\cite{sasaki2021behavioral}. Unlike our work that learns individual expertise levels of demonstrators, this work is mainly concerned with learning the best policy over noise-injected demonstrations without modeling the demonstrator identity or expertise. The second work (TRAIL)~\cite{yang2021trail} tackles this setting by leveraging suboptimal data to extract a latent action space, which is used alongside standard BC to train a policy on a small set of ``near-optimal'' expert demonstrations. In contrast, our approach does not rely on access to such near-optimal expert demonstrations. 
% Furthermore, since both works rely on BC, our method can be used on top of these approaches by utilizing the demonstrator identities. 
% (delete "recent" in first sentence if you want this line ^^)

%assume no access to expert data, whereas TRAIL is used to find an initialization from suboptimal data, but still relies on expert data for learning the final policy.

\noindent \textbf{Unsupervised Estimation of Expertise.}
The problem of inferring ground truth labels from crowd-sourced human data has been studied in biostatistics, education, and more recently computer vision, and NLP. These works generally tackle the problem using the Expectation Maximization (EM) algorithm to solve for the individual error rates of the human annotators~\cite{dawid1979maximum}. Furthermore, similar approaches have been applied to the crowd sourcing problem of labeling large image datasets~\cite{whitehill2009whose,raykar2010learning,welinder2010multidimensional}, learning a model over annotators to generate more accurate estimates. Specifically, one paper models each annotator and task using multidimensional variables representing difficulty, competence, expertise, and bias~\cite{welinder2010multidimensional}. Inspired by this, we apply a similar formulation to IL, addressing several challenges that go beyond the scope of supervised learning. More precisely, in the image domain, one can collect multiple annotator labels for many images and compare them to the ground truth. In our imitation learning setting, on the other hand, demonstrators may not visit the same states, states are intertwined through dynamics, and the optimal policy may give action distributions instead of a single optimal action.
%Due to the differences between image recognition tasks and IL, we only use prior work as inspiration, designing our own models as needed.\par

\section{Joint Estimation of Policy and Expertise} \label{sec: Model}
In this section we describe a joint model that learns from a dataset consisting of a mixture of demonstrations from demonstrators with varying, but unknown, levels of expertise. Our model both infers state-dependent expertise levels of the different demonstrators in the dataset, and recovers a single policy learned from all the demonstrations. 
% Accurate estimates of expertise levels help the model learn from heterogeneous demonstrations, leading to a better policy. In turn, a better estimate of the policy allows the model to more accurately predict the expertise levels.

\subsection{Problem Setup}
% We proceed by describing the dataset setup, the model of demonstrators, and the learning framework, and end by analyzing the recoverability of the optimal policy given suboptimal demonstrations under our model.

%We use a standard MDP $\MDP=\langle\States,\Actions,P,r,\gamma\rangle$ to define our problem.
\noindent \textbf{Dataset.} We collect a set \(\demons_i\) of trajectories $\tau=(s_0,a_0,\ldots,s_{t},a_{t})$ of varying length from each demonstrator \(i\).
We assume the trajectories from demonstrator \(i\) are sampled from some fixed underlying policy \(\pi_i\).
The full dataset \(\demons = \{ (i, \demons_i) \}_{i=1}^{m}\) is the union of the dataset from each of the \(m\) demonstrators, labeled by the demonstrator index. Generally speaking, each trajectory could exhibit close to random behavior, but could also come from a demonstrator with high expertise. We would like a model that identifies when demonstrations are suboptimal, to better learn a single policy from the mixed bag of demonstrations. 

%Our goal is to estimate the expertise of these demonstrations, to better learn a single policy from the mixed bag of demonstrations. 
%To do so, we need a model that identifies when demonstrations are suboptimal.
%We proceed by defining this model of demonstrators, describing the learning framework, and analyzing the recoverability of the optimal policy given suboptimal demonstrations from multiple demonstrators under our model.

\noindent \textbf{Demonstrator Model.} Our demonstrator model should be able to express state-dependent expertise. For example, demonstrator A may be adept at washing the dishes, while demonstrator B may be adept at vacuuming the floor, and modeling this state-dependent expertise can allow us to recognize and combine their strengths in different states.
To model such suboptimal policies, we will define two main components: 1) the expertise level of a demonstrator \textit{at a given state} and 2) the demonstrator's action distribution at a state as a function of their expertise level and the optimal action distribution at that state.

\noindent \textbf{\emph{1) Expertise Levels.}} 
%In general, even for the same demonstrator, we should be able to express different expertise levels at different states. For example, demonstrator A may be adept at washing the dishes, while demonstrator B may be adept at vacuuming the floor, and modeling this state-dependent expertise can allow us to recognize and combine their strengths in different states.
%
Drawing inspiration from annotator models~\cite{welinder2010multidimensional}, we model expertise levels using two embeddings: a $\dimembed$--dimensional state embedding using a deterministic map $\fembed:\States\rightarrow\reals^\dimembed$ (parameterized by $\phi$) from states to embeddings, and demonstrator embeddings \(\bomega \in \reals^{\numexperts\times\dimembed}\), where $\omega_i$ is a $\dimembed$--dimensional vector capturing the aptitude of demonstrator \(i\). Using these embeddings, we quantify the expertise level of demonstrator \(i\) at state $s$ as: 
\begin{align}
    \rho_\phi(s, \omega_i) = \skill(\langle\fembed(s), \omega_i \rangle),
    \label{eq:expertise}
\end{align}
where $\sigma:\reals\rightarrow(0,1)$ is the sigmoid function and \(\langle\cdot,\cdot\rangle\) denotes the inner product.

We can interpret each dimension of the embedding vector $\fembed(s)$ as a weighting of how relevant a latent skill is in acting correctly at that state $s$. A demonstrator's skill set is the $\dimembed$-dimensional embedding $\omega_i$ that expresses how adept the demonstrator is at each skill. This way we can measure how qualified demonstrator $i$ is at \textit{acting in} state $s$ by computing the inner product $\langle \fembed(s),\omega_i \rangle$ between the task encoding of the state, and the demonstrator's skill set $\omega_i$. 

\noindent \textbf{\emph{2) Demonstrator's Action Distribution.}} 
We now define the demonstrator's suboptimal policy as a function of their expertise level and the optimal policy \(\pi_{\theta^\star}\). 

We would like the expertise level \(0 \leq \rho_\phi(s, \omega_i) \leq 1\) of demonstrator \(i\) at state $s$ to be correlated with how close their action distribution is to the true action distribution \(\pi_{\theta^\star}(a|s)\), with \(\rho_\phi(s, \omega_i)=1\) corresponding to exactly \(\pi_{\theta^\star}(a|s)\) and \(\rho_\phi(s, \omega_i)=0 \) corresponding to a uniformly random distribution. We can satisfy this desiderata, separately for discrete and continuous action spaces, using the following models.

\noindent \textit{Discrete Action Space.}
When the action space is discrete, we use \(\rho_\phi(s, \omega_i)\) to interpolate between the optimal policy and the uniformly random policy which assigns probability \(1/|\Actions|\) to each action.
\begin{equation}\label{eq:annotator_discrete_1} 
\resizebox{.85\linewidth}{!}  % Hack to make it fit on one line
{%   
    $\pi(a|s,\omega_i, \phi, \pi_{\theta^\star}) = \, \rho_\phi(s, \omega_i) \pi_{\theta^\star}(a | s)  + 
	 \frac{1-\rho_\phi(s, \omega_i)}{|\Actions|}$%
}%
\end{equation}%
% \begin{align}\label{eq:annotator_discrete_1} 
% 	\pi(a|s,\omega, \phi, \pi_{\theta^\star}) =& \, \rho_\phi(s, \omega) \pi_{\theta^\star}(a | s)  + 
% 	 \frac{1-\rho_\phi(s, \omega_i)}{|\Actions|}
% % 	& \begin{cases}
% % 		\rho_{\phi, i, j} & \textrm{if } a_{i,j}=\pi_\theta(s_{ij}),\\
% % 		\frac{1-\rho_{\phi,i,j}}{|\Actions|-1}& \textrm{if } a_{i,j}\neq\pi_\theta(s_{ij}),
% % 	\end{cases}
% \end{align}
%
\noindent \textit{Continuous Action Space.}
For continuous action spaces, we will focus on Gaussian Mixture Model (GMM) action distributions, as in~\citet{robomimic2021}. Specifically, at a given state $s$ the optimal policy $\pi_{\theta^\star}(a|s)$ outputs a probability distribution over actions $a\in\Actions$ in the form of a GMM with \(k\) mixtures $\pi_{\theta^\star}(a|s) = \sum_{j=1}^{k} \alpha_j \mathcal{N}(a ; \mu^\star_j(s), \sigma^\star_j(s))$.
% \begin{equation}\label{eq:GMM_policy}
%     \pi_{\theta^\star}(a|s) = \sum_{j=1}^{k} \alpha_j \mathcal{N}(a ; \mu^\star_j(s), \sigma^\star_j(s))
% \end{equation}
Then, given a demonstrator with expertise level \(\rho_\phi(s,\omega_i)\) at state \(s\), we simply scale the variance of the optimal policy's GMM (equally for each component) by \(1/\rho_\phi(s,\omega_i)\). Thus, the probability of the demonstrator actions $\pi(a | s, \omega_i, \phi)$ modifies \(\pi_{\theta^\star}\) as follows:
\begin{equation}\label{eq:annotator_continuous}
\resizebox{.9\linewidth}{!}  % Hack to make it fit on one line
{%
    $\pi(a | s, \omega_i, \phi, \pi_{\theta^\star}) = \sum_{j=1}^{k} \alpha_j \mathcal{N}(a ; \mu^\star_j(s), \sigma^\star_j(s) / \rho_\phi(s,\omega_i))$
}%
\end{equation}

An expertise level of \(1\) corresponds to an expert whose recommendations align with the optimal policy, whereas an expertise level approaching \(0\) corresponds to an expert with close to uniformly random actions. More complex models of expertise can be explored, but we find that our simple single-valued expertise model already gives good improvements.

\subsection{Learning the Optimal Policy and Expertise Levels}

So far we have defined the model of demonstrators with respect to some optimal policy \(\pi_{\theta^\star}(a|s)\), i.e., a demonstrator's expertise level correlates with how close their policy is to \(\pi_{\theta^\star}(a|s)\). However, we do not have access to \(\pi_{\theta^\star}(a|s)\). 
Hence, we will define a parametric family of policies \(\{\pi_\theta : \theta \in \Theta\}\), and try to recover \(\pi_{\theta^\star}(a|s)\). For the analysis in the rest of the section we will assume that \(\Theta\) is well-specified (i.e. \(\theta^\star \in \Theta\)), and that all demonstrators explore all states with non-zero probability. Put together, we will be jointly learning \(\theta, \phi, \bomega\): the optimal policy, the state embedding network, and the demonstrator embeddings. Using a maximum likelihood approach, we optimize these variables with the loss corresponding to the negative log-likelihood (NLL) of data.
\begin{align}\label{eq:log_loss} 
    \loss(\theta,\phi,\bomega) &= - \mathbb{E}_{i, (s,a)} \Big[ \log \pi(a | s, \omega_i, \phi, \pi_\theta) \Big] \\
    &\approx - \frac{1}{|\demons|}\sum_{i}\sum_{(s,a)\in\demons_i}	\log  \pi(a | s, \omega_i, \phi, \pi_\theta)
\end{align}

\begin{figure}[!t]
    \centering
    
    \begin{tikzpicture}
    \begin{scope}[
        %on grid, 
        %latent/.append
        obs/.append style={minimum size=0.8cm},
        latent/.append style={minimum size=0.8cm},
        det/.append style={minimum size=0.8cm},
        auto
        ]
        
        \clip (-1,0.5) rectangle + (7,-3.5);

        % check borders of clip        
        % \node [box] at (-1,0.5) {a};
        % \node [box] at (6,-3) {a};
        
    	\node[obs] (s) {$s$}; %
    	\node[latent,right=of s] (omega) {$\omega_i$}; %
    	\node[latent,below=of s] (f) {$f_\phi$}; %
    	\node[det,below=of omega] (rho) {$\rho$}; %
    	
    	\node[latent,right=of omega] (pi) {$\pi_\theta$}; %
    	\node[det,below=of pi] (demonstrator) {$\pi$}; %
    	
    	\node[obs, right=of pi] (a) {$a$}; %
    	\node[det, below=of a] (loss) {$\loss$}; %
    	
    % 	\edge {s} {f};
    	\edge {s, f, omega} {rho};
    	\edge {rho, pi} {demonstrator};
    	\edge {demonstrator, a} {loss};
    	\draw[->] (s)  .. controls (-2.5,-2.5) and (2,-4) .. (loss);
    	
    	\path (rho) edge [->, >={triangle 45}, dashed,red,transform canvas={xshift=2mm}] node [swap] {$\omega$} (omega) ;%
    	\path (rho) edge [->, >={triangle 45}, dashed,red,transform canvas={yshift=-2mm}] node {$\phi$} (f) ;%
    	\edge[dashed,red,transform canvas={yshift=-2mm}] {demonstrator} {rho};
    	\path (demonstrator) edge [->, >={triangle 45}, dashed,red,transform canvas={xshift=2mm}] node [swap] {$\theta$} (pi) ;%
    	\edge[dashed,red,transform canvas={yshift=-2mm}] {loss} {demonstrator};
    
    \end{scope}
    \end{tikzpicture}
    \caption{We observe the state, action, and demonstrator index \(i\). Diamond nodes are deterministically computed. From the state embedding \(f_\phi(s)\) and the demonstrator embedding \(\omega_i\), we compute the expertise level \(\rho\). We combine \(\rho\) with the estimate \(\pi_\theta\) of the optimal policy to obtain the demonstrator policy \(\pi\). The loss \(\loss\) (Eq.~\ref{eq:log_loss})
    of \(\pi\) on state \(s\) and action \(a\) 
    is back-propagated to update the parameters \(\theta, \phi, \omega\).}
    \label{fig:model}
\end{figure}
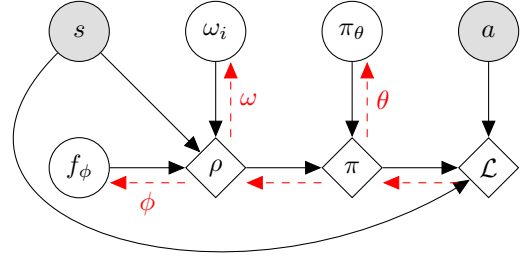
Although we can simply rely on the loss in Eq.~\ref{eq:log_loss} to learn the state embedding $\fembed$ (used in Eq.~\ref{eq:expertise}), it can be beneficial to consider the dynamics of the MDP environment as well, which may reveal more about the difficulty of each state. One popular approach is the DeepMDP framework~\cite{gelada2019deepmdp}, which uses an auxiliary loss to predict the environment dynamics in latent space. At a high level, this process uses the trajectories in our dataset as samples of the MDP dynamics to help learn a better state embedding $\fembed$. The details of this framework is described in Appendix~\ref{sec:appendix_deepmdp}.
% Because we only utilize this component in one of our experiments, we leave the details to the Appendix.

The overall learning framework can be seen in Fig.~\ref{fig:model}, where for demonstrator \(i\), we compute an expertise level \(\rho_\phi(s, \omega_i)\), that is then combined with 
our estimate \(\pi_\theta\) of the optimal policy to derive the demonstrator policy \(\pi(a | s, \omega_i, \phi, \pi_\theta)\) in Eq.~\ref{eq:log_loss}. We update all the parameters (\(\theta, \phi, \omega\)) to optimize our loss function.

To see why our loss function is suitable, we can rewrite the joint optimization equivalently only over \(\theta\), and show that the following objective serves as a \textit{proper} loss function.
\begin{align}\label{eq:log_loss_theta} 
    \loss(\theta) &= - \max_{\phi, \bomega} \mathbb{E}_{i,(s,a)} \Big[ \log \pi(a | s, \omega_i, \phi, \pi_\theta) \Big]
\end{align}
In other words, \(\theta^\star\) is the (non-unique) minimizer of \(\loss(\theta)\).
\begin{proposition} \label{prop:proper}
\(\loss(\theta)\) is a \textbf{proper} loss function.
\end{proposition}

In addition, it is easy to see that our framework generalizes standard behavioral cloning.
If we set the embedding vectors to large positive values everywhere, then the expertise levels \(\rho_\phi(s, \omega_i)\) for all states and all demonstrators will approach \(1\), in which case \(\loss(\theta)\) approaches \(\loss_{BC}(\theta)\).

\begin{remark}
ILEED recovers the standard behavioral cloning framework by setting all expertise levels to \(1\).
\end{remark}

Moreover, we will recover a different policy than behavioral cloning unless all demonstrators have identical policies. 
\begin{proposition} \label{prop:bc}
We have that \(\min_\theta \loss_{BC}(\theta) > \min_\theta \loss(\theta) \) unless all demonstrators have identical policies.
\end{proposition}

Hence, in the presence of different demonstrators, our framework will incorporate their varying expertise levels into its estimate of the optimal policy.

We have shown several nice properties of our loss function: 1) \(\theta^\star\) minimizes \(\loss(\theta)\), and 2) \(\loss(\theta, \phi, \bomega)\) generalizes \(\loss_{BC}(\theta)\), and 3) our framework incorporates varying expertise levels. However, we have not shown that \(\loss(\theta)\) is \textit{strictly proper}, i.e., \(\theta^\star\) may not be the unique minimizer and therefore it is unclear if we will recover the optimal policy.
To this end, we show that if we have knowledge of the state embedding \(f_\phi(s)\), then under some assumptions, we can uniquely recover the optimal policy \(\pi_{\theta^\star}\).

\begin{lemma} \label{lem:strictlyproper}
Let \(\phi\) be the ground truth state embedding parameters, and let \(\loss_{\phi}(\theta)\) be the loss in Eq.~\ref{eq:log_loss_theta} using fixed \(\phi\).
\[
    \loss(\theta) = - \max_{\bomega} \mathbb{E}_{i,(s,a)} \Big[ \log \pi(a | s, \omega_i, \phi, \pi_\theta) \Big]
\]
%and let \(\mathcal{C}\) be the family of functions \(C(s): S \rightarrow \mathbb{R}\) that maps states to reals such that \(\exists s : C(s) \neq 1\).
Under both the discrete and continuous action model (Eq.~\ref{eq:annotator_discrete_1} \& \ref{eq:annotator_continuous}), \(\loss_{\phi}(\theta)\) is a {\bf strictly proper} loss function if 
%
%for any \(C \in \mathcal{C}\), there exists a state 
%
for all states \(s_0\), there exists a set of other states \(s_{1:r}\) such that%
\begin{align} \label{eq:linear_dependent_condition}
   f_\phi(s_0) = \alpha_1 f_\phi(s_1) + \alpha_2 f_\phi(s_2) + \ldots + \alpha_r f_\phi(s_r)
\end{align}%

and no set of constants \(C_{0:r}\), with \(C_0 \neq 1\), satisfies the following conditions.
\(\forall i \in \{1,\ldots,m\}\): % and \(k \in \{1,\ldots,m\}\)
\begin{align*}
\sigma^{-1} ( \rho_\phi(s_0,\omega_i) / C_0 ) &= \sum_{k=1}^{r} {\alpha_k  \sigma^{-1} ( \rho_\phi(s_k,\omega_i)) / C_k ). }
\end{align*}
%

% and no set of constants \(C\) satisfies both following conditions:
% \[
% C_0 = \alpha_1 C_1 + \alpha_2 C_2 + \ldots + \alpha_m C_m
% \]
% and for all \(i,j \in \{1,\ldots,D\}\) and \(k \in \{1,\ldots,m\}\)
% \begin{align*}
%     %& \forall i,j \in \{1,\ldots,D\}, k \in \{1,\ldots,m\}\\
%     \frac{\sigma^{-1}(\langle f_\phi(s_k), w_i \rangle + C_k)}{\sigma^{-1}(\langle f_\phi(s_k), w_i \rangle)} 
%     = \frac{\sigma^{-1}(\langle f_\phi(s_k), w_j \rangle + C_k)}{\sigma^{-1}(\langle f_\phi(s_k), w_j\rangle)}
% \end{align*}
%
%\(f_\phi(s)\) can be written as a linear combination of the other embedding vectors \(\{f_\phi(s') : s' \neq s\}\) ... and \(\rho_\phi(s, \omega)\) ... and 
\end{lemma}

Intuitively, the challenge in uniquely recovering the optimal policy is that the true action distribution \(\pi_{\theta^\star}(\cdot|s)\) at a state may be expressed as a low-expertise version of another distribution \(\tilde{\pi}(\cdot|s)\). Our model might incorrectly recover \(\tilde{\pi}(\cdot|s)\), and compensate by decreasing the expertise levels of all demonstrators in a precise way. If all the state embeddings are linearly independent (e.g. basis vectors in high-dimensions), then our model can fully control the demonstrator expertise levels \(\rho_\phi(s, \omega_i)\) by tweaking the demonstrator embeddings \(\bomega\). On the other hand, the above result says that if the state embeddings are intertwined (Eq.~\ref{eq:linear_dependent_condition}), then the model cannot fully control \(\rho_\phi(s, \omega_i)\). Therefore the model cannot mistake \(\pi_{\theta^\star}(\cdot|s)\) for \(\tilde{\pi}(\cdot|s)\) since it cannot compensate for the different expertise levels. We include the proofs of these results in Appendix~\ref{sec:appendix_proofs}.

%To get intuition on the conditions in Lemma~\ref{lem:strictlyproper}, 
Moreover, note that the constraints on \(C_{0:r}\) are less likely to hold as the number of demonstrators $\numexperts$ grows.
In other words, assuming we have learned the correct state embeddings,
%(which we discuss in more detail next)
if the embeddings are intertwined enough then \ileed~can recover the optimal policy. On the other hand, BC will be unable to recover the optimal policy in the presence of suboptimality. We include a concrete example in Appendix~\ref{sec:appendix_concrete_example} showing the improvement of our method in the presence of demonstrators with varying optimalities.
%how our method improves upon behavioral cloning when taking into account demonstrator expertise levels.

% Finally, we conclude this section with a discussion on learning better state embeddings.
% We integrate an unsupervised learning auxiliary loss, which works by treating the dataset trajectories as samples of the transition dynamics.

% \noindent \textbf{Learning State Embeddings from Transitions.}
% Although we can simply rely on the NLL loss defined in Eq.~\ref{eq:log_loss} to learn the state embedding $\fembed$ (used in Eq.~\ref{eq:expertise}), it can be beneficial to consider the dynamics of the MDP environment as well, which may reveal more about the difficulty of each state. One popular approach is the DeepMDP framework~\cite{gelada2019deepmdp}, which uses an auxiliary loss to predict the environment dynamics in latent space. At a high level, this process uses the trajectories in our dataset as samples of the MDP dynamics to help learn a better state embedding $\fembed$. Because we only utilize this component in one of our experiments, we leave the details to the Appendix.

\noindent \textbf{Summary.} We defined a model of suboptimal demonstrators, where the action distribution of demonstrator \(i\) at a state \(s\) is determined by the state embedding $\fembed(s)$, the expertise embedding \(\omega_i\), and the optimal action distribution \(\pi_{\theta^\star}(a|s)\) (Eq.~\ref{eq:annotator_discrete_1},~\ref{eq:annotator_continuous}). Our method generalizes the standard BC framework to the case of demonstrators with varying expertise levels, enabling us to better handle demonstrators with varying state-dependent optimalities. Finally, we show the recoverability of the optimal policy when we have knowledge of the state embeddings, and integrate unsupervised techniques for learning these state embeddings.

\begin{figure*}
\subfloat[\label{fig:env_mini}]
[(i) \textit{Empty} - The agent starts at a random position. The objective is to reach the green square.\\
(ii) \textit{Lava} - The agent must pass through a narrow gap in a vertical strip of lava to. Touching the lava terminates the episode with a zero reward, making this environment useful for studying safe exploration.\\
(iii) \textit{Obstacles} - A large penalty is subtracted if the agent collides with an obstacle and the episode finishes, resulting in a possible negative reward.\\
(iv) \textit{Unlock} - The agent must unlock the door by first picking up a key and then entering the door.]
{\includegraphics[width=.38667\linewidth]{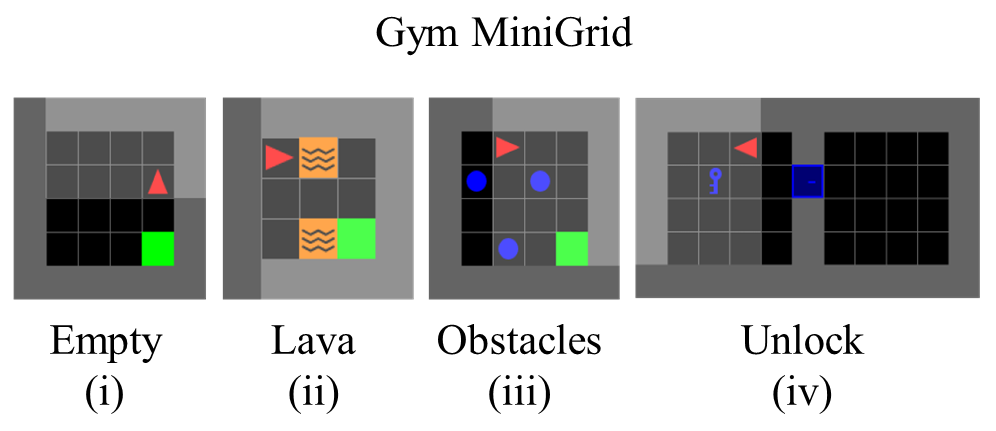}}\hfill
\subfloat[\label{fig:env_robo}]
[We utilized the low dimension Multi-Human (MH) Robomimic dataset which contains $50$ demonstrations from $6$ humans, split up into three groups based on their proficiency. We used the Square environment, a continuous control task in which a robotic arm must pick up a square nut and place it on a rod, and a binary reward is awarded for completing the task in the allotted time. The images shown were directly taken from the study~\cite{robomimic2021}.]
{\includegraphics[width=.29\linewidth]{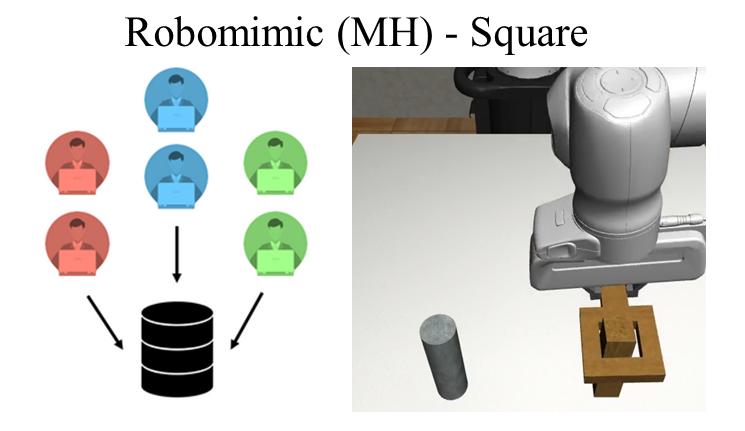}}\hfill
\subfloat[\label{fig:env_}]
[We examine games from the lichess database~\cite{mcilroy2020aligning}, consisting of $\sim$10M games in 2019 from players with rating ranging from 1k to 2k. We limit ourselves to endgame positions with only kings and pawns. We encode the state into a discrete vector and use a 3-layer MLP policy network and \(2\)-dimensional state and expertise embeddings. We divide the dataset into $5$ bins based on rating percentile (e.g. bottom \(20\%\) of players in the first bin).]
{\includegraphics[width=.29\linewidth]{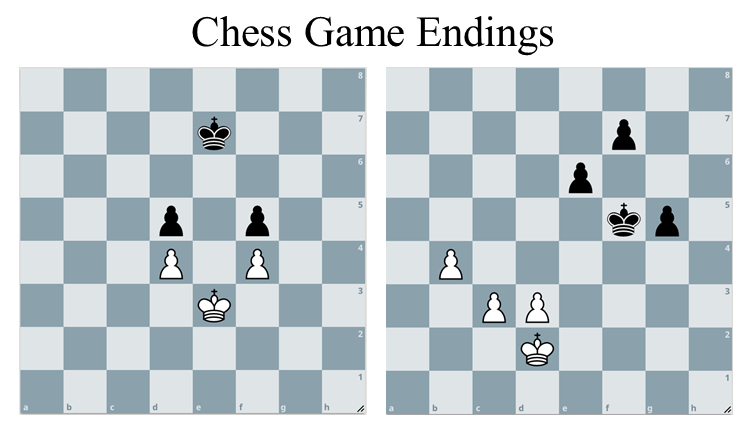}}
\caption{Environments used throughout our work: MiniGrid, Robomimic, and chess.}
\label{fig:env}
\end{figure*}

\section{Experiments} \label{sec: Results}
We will test if our algorithm can: (1) learn a policy from a mixture of \textit{simulated} demonstrations with varying levels of proficiency, (2) learn a policy from a mixture of \textit{human} demonstrations with varying levels of proficiency, (3) recover \textit{human} expertise levels even when learning an optimal policy is too challenging, and (4) learn a policy for multiple skills from a mixture of \textit{simulated} demonstrations with state-dependent noise.\par 

Note that we can use ILEED to learn either state-\textit{dependent} expertise levels, or state-\textit{independent} expertise levels where we assume a demonstrator has a single expertise value that is the same at all states. Out of the four aforementioned experiments in the paper, only the last experiment learns state-dependent expertise levels. For the first three experiments, we found that adding state-dependence did not improve performance. For the fourth experiment, state-dependent expertise was important due to the multi-skill nature of the task, and a state-independent approach did poorly. We show this by including an ablation along with the last experiment that studies the individual effect of state embeddings and demonstrator identities on ILEED's performance. Before going over our results, we briefly detail the specific environments and datasets used in our experiments.\par 
\subsection{Environments and Datasets}
For the first and last experiment we rely on simulated data, using $4$ MiniGrid~\cite{gym_minigrid} environments along with pre-trained policies with various levels of injected noise. All $4$ environments use a partially observable view with 3 input values per visible grid cell, and a maximum reward of one is given if the objective is reached with a small penalty subtracted for the number of steps to reach the goal. The environments are depicted in Fig.~\ref{fig:env}(a), along with a brief description for each. For the second experiment, which relies on suboptimal human data, we use the Robomimic dataset and codebase \cite{robomimic2021} which consists of various continuous control robotics environments along with corresponding sets of suboptimal human data. We depict the environment used and briefly describe the dataset in Fig.~\ref{fig:env}(b). For the third experiment we derive player rankings using human chess game-ending data provided by the lichess database~\cite{mcilroy2020aligning}, which is briefly explain in Fig.~\ref{fig:env}(c).

\subsection{Baselines}
Throughout the experiments we compare our model with $3$ other IL algorithms (BC, BC-RNN, GAIL), as well as one recent offline learning algorithms IRIS~\cite{mandlekar2020iris}. To the best of our knowledge, there are no IL algorithms which show good performance on suboptimal human datasets besides BC-based approaches. Current approaches that tackle this setting either rely on the reward signal \cite{fujimoto2019off,kumar2020conservative,mandlekar2020iris} (BCQ, CQL, IRIS), or break the offline assumption by using environment simulations \cite{ho2016generative,brown2019extrapolating,brown2020better,chen2020learning,fu2017learning} (GAIL, D-REX, T-REX, SSRR, AIRL). In addition, recent work has shown that BC-based approaches perform better compared to other offline learning techniques in settings with suboptimal demonstrations \cite{robomimic2021,florence2021implicit}. We thus treat BC-based approaches as our main baselines for the simulated experiments. To further test our model against the aforementioned recent algorithms, we directly compare with the results from Robomimic.

\subsection{Learning from Simulated Data}\label{sec:results_mini}
We first study how our algorithm performs on simulated suboptimal data, where we vary the optimality level by injecting noise into pre-trained policies. When simulating data, we use a set of $\numexperts$ state-independent expertise levels $\beta_i$ for $i\in\{1,\ldots, \numexperts\}$, and collect a fixed number of state-action pairs from each. Specifically $\beta_i=\rho_\phi(s, \omega_i) ~ \forall s\in \States$, where we use the discrete action space model defined in Eq.~\ref{eq:annotator_discrete_1} to interpolate between a random policy $\beta=0$ and the pre-trained policy $\beta=1$. All of the MiniGrid experiments use fully-connected neural networks with the Adam optimizer, with specific parameters left to Appendix~\ref{sec:appendix_implementation_details}. For the four aforementioned environments, the respective performance of the pre-trained policies along with their noised version are listed in Table~\ref{tab:pre_trained} in Appendix~\ref{sec:appendix_additional_experiments}.

Before moving on to our first experiment, we tested the relationship between the learned policy's performance and the corresponding NLL defined in Eq.~\ref{eq:log_loss}. By varying the number of restarts and choosing the policy with the highest likelihood, we can empirically test if our defined loss in Eq.~\ref{eq:log_loss} can also serve as a good validation metric for the final policy's performance. We display this in Table~\ref{tab:p_chart} of Appendix~\ref{sec:appendix_rew_vs_nll}, where we see that as the number of restarts increases, the policy's performance improves as well. Based on this insight, we set the number of restarts to $20$.
%, which is not computationally expensive given the small dataset.
As for the two baselines in this experiment, BC and GAIL, we also restart BC as many times as \ileed, choosing the policy with the lowest loss. In contrast, we only ran GAIL once, because unlike BC and \ileed, GAIL optimizes for a saddle point as opposed to a minimum, so we cannot use the lowest loss as a validation metric for choosing between restarts. 

We want to study how varying the population of demonstrators using simulated noise affects the performance of selected algorithms. To test this, we chose several sets of $\numexperts$ demonstrators, each with a different predefined set of expertise levels $\beta=\{\beta_1,
\ldots,\beta_M\}$. The four chosen sets are denoted as: $\beta$-\textit{1}:$\{\beta_1=0.99, \beta_{2:10}=0.01\}$, $\beta$-\textit{5}:$\{\beta_{1:5}=0.99, \beta_{6:10}=0.01\}$, $\beta$-\textit{10}:$\{\beta_{1:10}=0.99\}$, $\beta$-\textit{unif}:$\{\beta_i=0.05+0.1(i-1)\}$. By keeping the noise levels constant and re-sampling state-action pairs for each trial, we study how different distributions of expertise levels affect performance, showing this result in Table~\ref{tab:population}.\par

Our algorithm outperforms both BC and GAIL in all $4$ population settings when testing on the \textit{Empty} and \textit{Obstacles} MiniGrid environments.
%, where results for the other two are left to the Table~\ref{tab:population_extra} in Appendix~\ref{sec:appendix_minigrid}.
For the simpler \textit{Empty} environment, we see all algorithms were able to match the pre-trained policy's performance, though the BC struggled in the $\beta$-\textit{1} population shown on the first row, where only $1$ of the $10$ demonstrators is competent. In the more challenging \textit{Obstacles} environment, we see both GAIL and BC are unable to imitate the policy even when noise is diminished to $1\%$ in the $\beta$-\textit{10} population shown on the third row, where all $10$ demonstrators are fairly competent. Meanwhile \ileed~is able to achieve consistent performance even for the noisier populations. Overall GAIL showed inconsistent results and did not perform well in the \textit{Obstacles} environment, and we note again that GAIL also assumes access to environment dynamics. We include these comparisons for \textit{Lava} and \textit{Unlock} in Appendix~\ref{sec:appendix_minigrid}. Note that the rankings derived from the estimated expertise levels of the demonstrators correlate with the performance of the final policy as we show in Appendix~\ref{sec:appendix_minigrid_rank}.

\begin{table}[!t]
    \small
    \caption{Effect of Varying Population Expertise $\beta$}
    \raggedright We compute mean episodic reward of policies from different IL algorithms, over $20$ trials (for BC and \ileed).
    
    \label{tab:population}
    \centering
    \begin{tabular}{cccc|ccc}
        \toprule
        \multicolumn{1}{c}{$\beta$} & \multicolumn{3}{c}{Empty} & \multicolumn{3}{c}{Obstacles}\\
        \cmidrule(r){1-1}\cmidrule(r){2-4}\cmidrule(r){5-7}
        & \textbf{BC} & \textbf{GAIL} & \textbf{\ileed}
        & \textbf{BC} & \textbf{GAIL} & \textbf{\ileed}\\
        \textit{1} & $0.81$ & $0.96$ & $\mathbf{0.97}$ & $0.18$ & $-0.82$ & $\mathbf{0.91}$\\
        \textit{5} & $0.97$ & $0.96$ & $0.97$ & $0.66$ & $-0.77$ & $\mathbf{0.94}$\\
        \textit{10} & $0.97$ & $0.96$ & $0.97$ & $0.63$ & $-0.01$ & $\mathbf{0.94}$\\
        \textit{unif} & $0.97$ & $0.96$ & $0.97$ & $0.80$ & $-0.84$ & $\mathbf{0.90}$\\
        \bottomrule
    \end{tabular}
\end{table}
% We choose to include these results as they are interesting, but note that because GAIL utilizes environment dynamics to sample trajectories while learning, it can not be considered as offline learning, and hence is an unfair comparison to BC and \ileed. \par
\subsection{Learning from Suboptimal Human Data}\label{sec:results_robo}
Next, we test our model on the continuous control task \textit{Square} depicted in Fig.~\ref{fig:env}(b), where the dataset provided contains suboptimal human demonstrations. Specifically, the demonstrations used consists of three subsets of $100$ demonstrations provided by: two ``better" quality operators, two ``okay" operators, and two ``worse" operators. Like the original study, we use different combinations of the smaller subsets to investigate how suboptimal human data affects performance. Using this dataset, we are able to compare our method with three IL algorithms (BC, BC-RNN, HBC), as well as three recent offline learning algorithms (BCQ, CQL, IRIS) which differ from the IL setup by also utilizing reward information. Nonetheless, we are able to outperform all six methods for every combination of the suboptimal dataset, and hence only include results for the strongest baselines of each group: BC-RNN and IRIS. This result is displayed in Table~\ref{tab:robomimic}, where we note that no restarts were used in our model for fair comparison with results reported from \cite{robomimic2021}.\par
We can see from the results that modeling human expertise levels as done by \ileed~provides consistent improvement over other IL algorithms.
%---which do better on human datasets compared to offline RL algorithms. 
As noted by the cited study: BC-RNN is a strong baseline on suboptimal human data, but there is room for improvement. We see from Table~\ref{tab:robomimic} that \ileed~does in fact improve over BC-RNN as it is better at utilizing suboptimal demonstrations. 
%In particular, if we compare the results on the $100$ Better demonstrations and $100$ Okay demonstrations to the $200$ Worse-Better demonstrations and $200$ Worse-Okay demonstrations, we observe how adding $100$ Worse demonstrations impacts the performance of each algorithm. Though most algorithms (not shown here) decline in performance, our method is able to improve from the added data, surpassing even BC-RNN. 
Finally, we can see that \ileed~outperforms BC-RNN in all settings with an average $4.8\%$ increase in final reward. This shows that learning a model for demonstrator's expertise can significantly boost performance by taking more advantage of suboptimal data compared to methods that ignore demonstrator expertise.\par
\begin{table}[!t]
    \small
    \centering
    \captionsetup{justification=centering}
    \caption{Suboptimal Human Data for Continuous Control}
    \raggedright Our method outperforms all other methods in all settings. We copy results for the two strongest methods~\cite{robomimic2021}, and average \ileed~over $3$ trials as done in the Robomimic study.
    
    \label{tab:robomimic}
    \centering
    \begin{tabular}{cccc}
        \toprule
        \textbf{Dataset} & \textbf{BC-RNN} & \textbf{IRIS} & \textbf{\ileed} (ours)\\
        \midrule
        All & $\mathbf{78.0\pm4.3}$ & $52.7\pm5.0$ & $\mathbf{78.0\pm1.6}$\\
        \midrule
        Worse & $39.3\pm3.8$ & $38.7\pm0.9$ & $\mathbf{46.7\pm4.7}$\\
        Okay & $45.3\pm2.5$ & $42.0\pm3.3$ & $\mathbf{53.3\pm2.5}$\\
        Better & $66.0\pm2.8$ & $60.0\pm1.6$ & $\mathbf{72.7\pm3.8}$\\
        \midrule
        Worse-Okay & $55.3\pm0.9$ & $43.3\pm2.5$ & $\mathbf{59.3\pm3.8}$\\
        Worse-Better & $73.3\pm6.2$ & $56.7\pm3.4$ & $\mathbf{77.3\pm6.8}$\\
        Okay-Better & $74.0\pm2.8$ & $56.7\pm3.8$ & $\mathbf{77.3\pm0.9}$\\
        \midrule
        % Can comment out later
        Total & $61.6\pm3.3$ & $50.0\pm2.9$ & $\mathbf{66.4\pm3.4}$\\
        \bottomrule
    \end{tabular}
\end{table}
\subsection{Estimating Expertise from Human Data}\label{sec:results_chess}
For our second human experiment, we explore the ambitious task of learning to play chess endgames purely from data~\cite{mcilroy2020aligning}, without access to the environment (i.e., without knowing the rules of the game or accessing rewards). This task is extremely difficult, since most chess-playing agents assume access to the game's rules and rely on some form of self-play or tree-search to reach good performance~\cite{pascutto_leela_2018,romstad_stockfish_2021}. Nevertheless, this task serves as a good benchmark for validating our model's estimation of the expertise level of demonstrators. After splitting our dataset into $5$ bins based on rating percentile (Fig.~\ref{fig:env}(c)), we recover the expertise $\rho$ of the bins.
%\begin{table}[!t]
    %\centering
    %\captionsetup{justification=centering}
    %\caption{We group chess players into \(5\) bins based on rating percentile, and try to recover the expertise \(\rho\) of the bins.
    %Our estimate is monotonic w.r.t. the rating of the bins.}
    %\label{tab:chess}
    % \begin{tabular}{c|ccccc}
    % \toprule
    % Bins & 1 & 2 & 3 & 4 & 5\\
    % \midrule
    % Expertise & $.9207$&$.9274$&$.9298$&$.9328$&$.9329$\\
    % \bottomrule
    % \end{tabular}
%\end{table}
\begin{center}
    \begin{tabular}{c|ccccc}
    \toprule
    Bins & 1 & 2 & 3 & 4 & 5\\
    \midrule
    Expertise & $.9207$&$.9274$&$.9298$&$.9328$&$.9329$\\
    \bottomrule
    \end{tabular}
\end{center}
In the above table, we see that our predicted expertise levels (averaged over states in the dataset) of the bins align with the ground truth skill levels of the bins -- the higher the ground truth rating of the bins, the higher our estimated expertise level. This suggests that our framework can serve a dual purpose as an estimator of demonstrator expertise levels. Though these expertise values seem close in absolute terms, we note that a difference of a few percentage points in accuracy can lead to a large rating difference (cf. Fig. 3 in~\cite{mcilroy2020aligning}), and that the takeaway of our experiment is the monotonicity of our recovered expertise values.
Lastly, in Table~\ref{tab:chess_policy} (in Appendix~\ref{sec:appendix_chess_policy}), we check that our method also outperforms BC in terms of the performance of the learned policy, as measured by treating StockFish evaluations as the reward in the environment. Though we did not expect to recover a good policy, given the difficulty of the task, the improvement over BC may be interesting for further investigation.

\subsection{Learning from Expertise in Different Skills}\label{sec:results_multi}
So far, we have not used our model's state-dependent components, namely the state embedding $\fembed(s)$ and the auxiliary loss provided by the DeepMDP framework. In fact, we note that surprisingly, the simulated MiniGrid experiments performed in Section~\ref{sec:results_mini} produced near-identical results with and without the auxiliary loss. Thus before concluding, we study the state dependent components by simulating an environment with multiple skills, where each skill can be seen as achieving an independent task within the environment.

In this experiment, we utilize the environments \textit{Unlock}, \textit{Lava}, and \textit{Empty}. We note that these three environments exhibit the need for independent skills, i.e., a policy trained on one of them does not necessarily do well on the other two. The reason we focus on independent skills is to ensure training a policy on each and all task will lead to full coverage over all the desired states and skills; however, our method can also generalize to settings with non-independent skills as well.
% Though our method generalizes to non-independent skills, by restricting ourselves we make sure that in order to succeed the policy must learn all $3$ skills. 
We place the $3$ environments in succession, such that an agent must successfully perform all $3$ tasks to receive a reward of $1$. When collecting trajectories, we allow continuation to the next environment even if the policy failed in the current environment. For evaluation, we average the agents' performance on all environments. For the datasets we use $3$ demonstrators, one being an expert in each skill. Specifically, we collect $10000$ state action pairs from all $3$ demonstrators, where they act according to the optimal (pre-trained) policy in the environments they are \textit{skilled} in, and suboptimally in the other two environments. We control the level of suboptimality by $\beta$, which refers to the probability that the demonstrator acts optimally for the environments they are \textit{unskilled} in. For example, a demonstrator skilled in \textit{Unlock} with $\beta=0.1$ will always act according to the pre-trained policy when in the \textit{Unlock} environment, while only following the corresponding pre-trained policies $10\%$ of the time in the other two environments (acting randomly otherwise). This way, the noisiest dataset at $\beta=0$ still contains some optimal demonstrations for all $3$ environments, and as we increase $\beta$ to $1$, all demonstrations come from the corresponding pre-trained policy. We show the result for $5$ values of expertise $\beta$ in Table~\ref{tab:multi_minigrid}.\par
We can see that our model consistently outperforms the average demonstrator for $\beta<1$, and even the best demonstrator for $\beta<0.5$. This is expected for lower $\beta$, as even the best demonstrator can only be successful at one of the tasks due to the independence of tasks. We note that the optimal dataset at $\beta=1$ contains no noisy demonstrations, in which case our model is able to come within $5\%$ of the best demonstrator. As we decrease $\beta$, each demonstrator provides more suboptimal state-action pairs for $2$ of the $3$ environments, but our model is still able to combine their expertise and learn a policy which is skilled in multiple environments. We show trajectories in Fig.~\ref{fig:traj} of the Appendix, noting that for all settings our learned policy adequately performs on $2$ environments, failing mostly in \textit{Unlock}. Additionally, we provide an ablation study in Appendix~\ref{sec:appendix_ablation} to test the individual effect of state embeddings and demonstrator identities on \ileed's performance, showing that both components contribute to the high performance of \ileed~in this multi-skill experiment. 
% \begin{table}[!t]
%     \small
%     \centering
%     \captionsetup{justification=centering}
%     \caption{\centering Learning Multiple Skills from Suboptimal Data.\\
%     Mean and standard deviation of episodic reward of the best demonstrator policy and the policy learned via \ileed, computed over $100$ trials.}
%     \label{tab:multi_minigrid}
%     \begin{tabular}{ccc}
%     \toprule
%     $\beta$ & \textbf{Best Demonstrator} & \textbf{\ileed}\\
%     \midrule
%     $0.01$ & $0.44\pm0.04$ & $\mathbf{0.70\pm0.06}$\\
%     $0.10$ & $0.57\pm0.04$ & $\mathbf{0.80\pm0.06}$\\
%     $0.20$ & $0.67\pm0.04$ & $\mathbf{0.86\pm0.05}$\\
%     $0.50$ & $0.66\pm0.04$ & $\mathbf{0.83\pm0.06}$\\
%     $1.00$ & $0.93\pm0.01$ & $0.91\pm0.04$\\
%     \bottomrule
% \end{tabular}
% \end{table}
\begin{table}[!t]
    \caption{Learning Multiple Skills from Suboptimal Data}
    {\small \raggedright Mean and standard deviation of episodic reward (over $100$ trials) for all demonstrators, the best demonstrator and \ileed.}
    
    \label{tab:multi_minigrid}
    \centering
    \begin{tabular}{cccc}
        \toprule
        $\beta$ & \textbf{All Demons.} & \textbf{Best Demons.} & \textbf{\ileed}\\
        \midrule
        $0.01$ & $0.40\pm0.04$ & $0.44\pm0.04$ & $\mathbf{0.70\pm0.06}$\\
        $0.10$ & $0.52\pm0.05$ & $0.58\pm0.05$ & $\mathbf{0.80\pm0.06}$\\
        $0.20$ & $0.67\pm0.04$ & $0.76\pm0.04$ & $\mathbf{0.86\pm0.05}$\\
        $0.50$ & $0.85\pm0.03$ & $0.90\pm0.01$ & $0.86\pm0.04$\\
        $1.00$ & $0.94\pm0.01$ & $0.94\pm0.01$ & $0.89\pm0.04$\\
        \bottomrule
    \end{tabular}
\end{table}
\section{Discussion}
\noindent \textbf{Summary.} We present \ileed~-- an approach for IL from demonstrators with varying but unknown state-dependent expertise. By jointly optimizing for the optimal policy and demonstrator expertise, we learn a better policy as compared to imitation learning
baselines, and also are able to recover accurate estimates of the ground-truth expertise levels. 

Our framework highlights the important problem of IL from datasets with multiple demonstrators.
Datasets collected from different demonstrators are already prevalent across continuous control and discrete tasks~\cite{robomimic2021,mcilroy2020aligning}, and will only increase in relevance as the need for large and diverse datasets inevitably grows~\cite{sharma2018multiple}. We show that in these settings, unsupervised estimation of the demonstrator expertise gives a large boost in performance.

\noindent \textbf{Limitation and Future Work.}
Our work is limited in a number of ways. First, some of our model's theoretical properties and predictive power rely on recovering effective state embeddings, which can be challenging in practice, e.g., when the state space is not fully explored by the demonstrators.
% when demonstrators do not explore the same states.
% Furthermore, we do not spend extensive time analyzing how well our method approximates expertise levels, instead focusing on the quality of the actual policies recovered. 
%Moreover, our method (and imitation learning in general) still struggles in challenging domains such as chess or even MiniGrid \textit{Unlock}. 
Moreover, we have yet to analyze the relationship between recovered expertise values $\rho$ and demonstrators' true expertise levels for environments with state-dependent expertise. Finally, our method currently has a simple model of suboptimality that uses a single-valued expertise level at each state. More complex demonstrator models can be explored to capture different modes of suboptimality in demonstrators.
%leverages unsupervised learning of the expertise levels, the model might be susceptible to adversarial demonstrators that makes it assign low expertise to good demonstrators.
%
%since our method leverages nothing beyond the demonstrator identity of each trajectory, improvements in performance are a result of the inductive bias of our model. One interpretation of this inductive bias is \textit{outlier filtering}, where the final policy is less affected by demonstrators identified to have low expertise. A limitation of this inductive bias is that the model might be susceptible to adversarial noise, e.g., if the outliers are demonstrators with high expertise.

%\noindent \textbf{Future Work.}
Nevertheless, we believe our framework provides a novel and effective approach for addressing the prevalent problem of learning from demonstrators with varying levels of expertise. 
% estimating expertise of multiple demonstrators shows consistent improvement over imitation learning baselines. Many existing datasets and domains have information about the demonstrator identity, which our model exploits to improve the final learned policy. 
Our framework is general, and lends itself to many possible directions of future work. As part of future work, we plan to model the uncertainty over the expertise levels and consider different notions of suboptimality beyond noisy action distributions.
% For example, we can model the uncertainty of the expertise levels, or consider different notions of suboptimality beyond noisy action distributions.
%which improve robustness to adversarial demonstrators. 
%Additionally, estimating demonstrator expertise for inverse reinforcement learning, or in the presence of some reward signal (i.e. offline learning) may also be interesting directions.

% Restate that we are doing imitation learning (offline without reward), and have offline evaluation on top of this
% 

% Future Directions: (1) The model is general, hence you can easily create different demonstrator models (2) Possible to utilize reward signal as well (or rather use this idea in an offline RL setting),

% Limitations: (1) No robustness to adversarials in the dataset, (2) more tests with multi-skill environments, larger human datasets, (different demonstrator models can go here)

%\clearpage
\section*{Acknowledgements}
This research was supported by NSF (1941722, 2006388, 2125511, 2003035, 1952920, 1651565), ARO (W911NF2110125), AFOSR, ONR, and Ford. 
%\bibliography{refs}
%\bibliographystyle{icml2022}

%%%%%%%%%%%%%%%%%%%%%%%%%%%%%%%%%%%%%%%%%%%%%%%%%%%

\clearpage

\appendix
\section{Learning State Embeddings from Transitions} \label{sec:appendix_deepmdp}
Although we can simply rely on the NLL loss defined in Eq.~\ref{eq:log_loss} to learn the state embedding $\fembed$ (used in Eq.~\ref{eq:expertise}), it can be beneficial to consider the dynamics of the MDP environment as well, which may reveal more about the difficulty of each state. We do not have access to the MDP dynamics, but we can use the provided state transitions as samples from the dynamics to better learn state embeddings.

One popular approach is the DeepMDP framework~\cite{gelada2019deepmdp}, which attempts to predict the environment dynamics in latent space.
Using DeepMDP as an auxiliary task in the Atari 2600 domain has shown large performance improvements over model-free RL~\cite{gelada2019deepmdp}.
DeepMDP trains an embedding function by minimizing two losses: prediction of rewards and prediction of the distribution over next latent states.
In our case, we do not have access to the rewards, so instead we replace the DeepMDP reward loss with our log-likelihood loss in Eq.~\ref{eq:log_loss}.

First, we define a latent transition network $g_\psi:\reals^\dimembed\times\Actions\rightarrow\reals^\dimembed$ parametrized by $\psi$, which takes as input the current state embedding $\fembed(s)$ and the action $a$, and outputs the predicted next-state embedding. Then, we take a transition tuple \((s,a,s')\) and minimize the distance between the predicted next-state embedding $g_\psi(\fembed(s),a)$  and the true next-state embedding $\fembed(s')$ on a metric $D$, which we choose to be the smooth $L1$ metric.
\begin{equation}\label{eq: latent_loss}
\loss(\psi, \phi)= \mathbb{E}_{(s,a,s')}\Big[ D \big(g_\psi(\fembed(s),a),\fembed(s')\big) \Big]
\end{equation}
%where $D$ is chosen to be the smooth $\ell$-1 metric. 
We augment our loss in~\ref{eq:log_loss} with this auxiliary loss $\loss(\psi, \phi)$ to help us learn the parameters $\phi$ used for the state embedding $\fembed$.
Intuitively, $\loss(\psi, \phi)$ encourages the learned embeddings $\fembed$ to admit predictable transitions, which hopefully pushes the embeddings of similar states close together.
At a high level, this process uses the trajectories in our dataset as samples of the MDP dynamics to help learn a better state embedding $\fembed$.

\section{Concrete Example of Embedding Values} \label{sec:appendix_concrete_example}

We show a concrete example on a simple 3-state and 3-action task. First, we show the state embedding and the optimal action distribution for the three states, one per row.

% optimal policy, rows are states, columns are actions
{\bf State Embedding and Optimal Policy}\\
\begin{tabular}{|c|c|c|c|}
     \hline
     \(f_\phi(s)\) & \(a_1\) & \(a_2\) & \(a_3\)\\
     \hline
     \{1,0\} & .8 & .1 & .1 \\
     \hline
     \{1,1\} & .0 & .5 & .5 \\
     \hline
     \{0,1\} & .1 & .1 & .8 \\
     \hline
\end{tabular}

Next, we assume that we have two demonstrators, who are suboptimal w.r.t. the optimal action distribution. Note that demonstrator 1 is optimal in the first two states, and demonstrator 2 is optimal in the last two states.

% 
% expert at f[0]
{\bf Demonstrator 1 \quad\quad Demonstrator 2}\\
\begin{tabular}{|c|c|c|}
     \hline
     \(a_1\) & \(a_2\) & \(a_3\)\\
     \hline
     .8 & .1 & .1  \\
     \hline
     .0 & .5 & .5 \\
     \hline
     .3 & .3 & .4 \\
     \hline
\end{tabular}%
\quad\quad\,\,\,%
% expert at f[1]
\begin{tabular}{|c|c|c|}
     \hline
     \(a_1\) & \(a_2\) & \(a_3\)\\
     \hline
     .4 & .3 & .3  \\
     \hline
     .0 & .5 & .5 \\
     \hline
     .1 & .1 & .8 \\
     \hline
\end{tabular}

Given a large dataset of demonstrations from demonstrators 1 and 2, ILEED and BC can recover the following policies.

% \ileed
% optimal \rho = 3/7
% so sigma^-1(3/7) = -.287
{\bf \ileed (left) vs BC (right) recovery}\\
\begin{tabular}{|c|}
     \hline
     \(\omega_1\) \\
     \{50,-1.79\} \\
     \hline
     \(\omega_2\) \\
     \{-1.79,50\} \\
     \hline
\end{tabular}%
%\,
%\quad%
\begin{tabular}{|c|c|c|}
     \hline
     \(a_1\) & \(a_2\) & \(a_3\)\\
     \hline
     .8 & .1 & .1  \\
     \hline
     .0 & .5 & .5 \\
     \hline
     .1 & .1 & .8 \\
     \hline
\end{tabular}%
\quad%
% BC
\begin{tabular}{|c|c|c|}
     \hline
     \(a_1\) & \(a_2\) & \(a_3\)\\
     \hline
     .6 & .2 & .2  \\
     \hline
     .0 & .5 & .5 \\
     \hline
     .2 & .2 & .6 \\
     \hline
\end{tabular}

log-likelihood: {\ul {-0.807}} \quad \quad \quad \quad -0.865

ILEED can learn the demonstrators embeddings (e.g. demonstrator 1 is better at skill 1, and demonstrator 2 is better at skill 2), and account for their suboptimalities to recover the optimal action distribution in all three states. The model can recover this via maximum likelihood, since it gives a better log-likelihood than standard BC. On the other hand, BC will average the demonstrations, and act suboptimally in the first and third state. 

%Our model can learn the correct optimal policy, since it gives a better log-likelihood.

\section{Additional Experiments} \label{sec:appendix_additional_experiments}

\subsection{Relationship between reward and log-likelihood} \label{sec:appendix_rew_vs_nll}
First we test the relationship between the learned policy's performance and the corresponding NLL defined in Eq.~\ref{eq:log_loss}. We did this by varying the number of restarts, and evaluating the performance of the policy with the highest likelihood. We ran this over $100$ trials, each time collecting a new set of $1000$ state-action pairs from $10$ independent demonstrators with uniformly generated noise levels ranging $\alpha\sim U(0,0.5)$. From this we estimated two values, the probability of the policy exceeding the mean demonstrator performance denoted by $p$, and the probability of the policy exceeding the best demonstrator's performance denoted by $p^*$. The results are provided in Table~\ref{tab:p_chart}.

\begin{table}[h!]
    \centering
    \caption{Effect of Restarting}
    \raggedright Estimated values for $p$ and $p^*$ computed over $100$ trials.
    
    \label{tab:p_chart}
    \centering
    \begin{tabular}{cccc}
        \toprule
        \multicolumn{1}{c}{Environment} & \multicolumn{3}{c}{Num. of Restarts}\\
        \cmidrule(r){1-1}\cmidrule(r){2-4}
        & $1$ & $5$ & $20$\\
        \midrule
        \textbf{Empty} & $0.94,0.82$ & $1.00,1.00$ & $1.00,1.00$\\
        \textbf{Lava} & $0.79,0.63$ & $0.99,0.87$ & $1.00,0.95$\\
        \textbf{Obstacles} & $0.37,0.11$ & $0.77,0.31$ & $0.84,0.29$\\
        \textbf{Unlock} & $0.25,0.00$ & $0.48,0.00$ & $0.59,0.00$\\
        \bottomrule
    \end{tabular}
\end{table}

Note that because the pre-trained policy's performance on the Unlock environment is particularly sensitive to noise, as shown by Table~\ref{tab:pre_trained}, we expect our algorithm to do relatively worse on this environment. On top of this, we see the pre-trained policy ($\beta=1.0$) for \textit{Unlock} does slightly worse than the noisy policy with $\beta=0.9$. This is because the pre-trained MLP policy does not fully solve \textit{Unlock} as it has no history component, getting stuck at certain states, hence adding a small amount of noise helps the policy escape such states.  

\begin{table}[t!]
    \centering
    \captionsetup{justification=centering}
    \caption{Pre-trained Policy Performance}
    \raggedright Mean episodic reward computed over $1000$ runs.
    
    \label{tab:pre_trained}
    \centering
    \begin{tabular}{ccccc}
        \toprule
        \multicolumn{1}{c}{Environment} & \multicolumn{4}{c}{Expertise level $\beta$}\\
        \cmidrule(r){1-1}\cmidrule(r){2-5}
        & $1.0$ & $0.9$ & $0.5$ & $0.1$\\
        \midrule
        \textbf{Empty} & $0.97$ & $0.97$ & $0.90$ & $0.44$\\
        \textbf{Lava} & $0.95$ & $0.88$ & $0.67$ & $0.05$\\
        \textbf{Obstacles} & $0.95$ & $0.94$ & $0.86$ & $0.31$\\
        \textbf{Unlock} & $0.87$ & $0.90$ & $0.83$ & $0.27$\\
        \bottomrule
    \end{tabular}
\end{table}
As we increase the number of restarts, the policy's performance improves as well. Not only does this result show that our method is able to outperform the best demonstrator, it implies that our defined NLL is a good metric for evaluating the optimal policy. Based on this, we set the number of restarts to $20$, where we note that given the small sizes of the datasets we utilize, restarting does not drastically affect computation time. Following this, we compared how our algorithm performed with respect to other baselines.\par

\subsection{MiniGrid Lava and Unlock} \label{sec:appendix_minigrid}

We show the remainder of the results from Section~\ref{sec:results_mini} in Table~\ref{tab:population_extra}. Overall GAIL showed inconsistent results for the \textit{Unlock} environment, and performed poorly on \textit{Lava}. Overall, \ileed~was able to to take advantage of the noisy demonstrations, while BC suffered as $\beta$ decreased.
\begin{table}[h!]
    \small
    \caption{Effect of Varying Population Expertise $\beta$}
    \raggedright We compute mean episodic reward of policies from different IL algorithms, $20$ trials (for BC and \ileed).
    
    \label{tab:population_extra}
    \centering
    \begin{tabular}{cccc|ccc}
        \toprule
        \multicolumn{1}{c}{$\beta$} & \multicolumn{3}{c}{Lava} & \multicolumn{3}{c}{Unlock}\\
        \cmidrule(r){1-1}\cmidrule(r){2-4}\cmidrule(r){5-7}
        & \textbf{BC} & \textbf{GAIL} & \textbf{\ileed}
        & \textbf{BC} & \textbf{GAIL} & \textbf{\ileed}\\
        \textit{1} & $0.95$ & $0.00$ & $0.95$ & $0.15$ & $0.96$ & $0.57$\\
        \textit{5} & $0.95$ & $0.00$ & $0.95$ & $0.75$ & $0.18$ & $0.81$\\
        \textit{10} & $0.95$ & $0.07$ & $0.95$ & $0.49$ & $0.74$ & $0.79$\\
        \textit{unif} & $0.95$ & $0.00$ & $0.95$ & $0.79$ & $0.01$ & $0.78$\\
        \bottomrule
    \end{tabular}
\end{table}

% \begin{table}[!h]
%     \centering
%     \captionsetup{justification=centering}
%     \caption{Effect of Restarting.\\
%     Estimated values for $p$ and $p^*$ computed over $100$ trials.}
%     \label{tab:p_chart}
%     \begin{tabular}{cccc}
%     \toprule
%     \multicolumn{1}{c}{Environment} & \multicolumn{3}{c}{Num. of Restarts}\\
%     \cmidrule(r){1-1}\cmidrule(r){2-4}
%     & $1$ & $5$ & $20$\\
%     \midrule
%     \textbf{Empty} & $0.94,0.82$ & $1.00,1.00$ & $1.00,1.00$\\
%     \textbf{Lava} & $0.79,0.63$ & $0.99,0.87$ & $1.00,0.95$\\
%     \textbf{Obstacles} & $0.37,0.11$ & $0.77,0.31$ & $0.84,0.29$\\
%     \textbf{Unlock} & $0.25,0.00$ & $0.48,0.00$ & $0.59,0.00$\\
%     \bottomrule
% \end{tabular}
% \end{table}
\subsection{MiniGrid Recovered Rankings}
\label{sec:appendix_minigrid_rank}
\begin{table}[h!]
    \centering
    \captionsetup{justification=centering}
    \caption{Pre-trained Policy Performance}
    \raggedright Mean episodic reward computed over $1000$ runs.
    \label{tab:minigrid_rank}
    \centering
    \begin{tabular}{ccc}
        \toprule
        \multicolumn{3}{c}{Env: Obstacles}\\
        \midrule
        Range & \textbf{Reward} & \textbf{Rank Loss} \\
        \midrule
        low & $0.73$ & $0.27$ \\
        high & $0.86$ & $0.00$ \\
        % \midrule
        % \multicolumn{3}{c}{Env: Empty}\\
        % \midrule
        % low & $0.97$ & $0.40$ \\
        % high & $0.97$ & $0.00$ \\
        \bottomrule
    \end{tabular}
\end{table}
Our third experiment described in Section~\ref{sec:results_chess} of the paper showed that ILEED can learn to rank demonstrators. Here, we examine if a good ranking of the demonstrators correlates with good performance of the final policy. Using grid-world \textit{Obstacles} as setup in Section~\ref{sec:results_mini}, we designed two demonstrator populations with the same average expertise, but one has low range ($0.15$ to $0.85$) (harder to rank) and one has high range ($0.05$ to $0.95$) (easier to rank). Table~\ref{tab:minigrid_rank} suggests that lower ranking loss relates to better reward, which is interesting since the average expertise of the two populations are identical.

\subsection{Chess policy} \label{sec:appendix_chess_policy}

In Table~\ref{tab:chess_policy} we show results for the chess-playing policy learned via imitation learning on the lichess database. As mentioned before, learning to play chess without knowing the rules of the game is extremely challenging, since we cannot improve via self-play. As a result, both BC and \ileed learn relatively poor policies (though still better than random). To interpret the results, note that we took the chess endgame positions from the database and polled a chess move from the learned policies. Then, we measured the difference in StockFish evaluation (ran for \(2\) seconds for each position) between the starting and ending positions. We set the evaluation of positions with inevitable mate to \(-100\). One very crude way to interpret the results is that \ileed~will blunder into an inevitable mate around \(3\%\) of the time, compared to \(4\%\) for BC and \(12\%\) for random. We note that this is not the most precise interpretation, since it ignores the change in evaluation of non-mating positions.
\begin{table}[h!]
    \centering
    \caption{Chess Policy Evaluation: average pawn loss per move.
    %\\Optimal (human): 0.25 \(\pm\) 0.11. Random: -12.30 \(\pm\) 0.11
    }
    \label{tab:chess_policy}
    \begin{tabular}{ccc}
        \toprule
        \textbf{Random} & \textbf{BC} & \textbf{\ileed} \\
        \midrule
        \(-12.30 \pm 0.11\) & \(-4.00 \pm 0.11\) & \(\mathbf{-3.27 \pm 0.09}\) \\
        \bottomrule
    \end{tabular}
\end{table}

\subsection{Ablation Studies}
\label{sec:appendix_ablation}
Lastly, we show our ablation studies discussed in Section~\ref{sec:results_multi}. Using the same experimental setup as Table~\ref{tab:multi_minigrid}, we conducted an additional study to test the individual effect of state embeddings and demonstrator identities on \ileed's performance. In Table~\ref{tab:ablations}, SInd is the state-independent variant of \ileed~that does not use the state embeddings, while DInd is the demonstrator-independent variant of \ileed~that uses state embeddings but removes the demonstrator identities treating them uniformly. As shown in the table, for varying levels of $\beta$ (demonstrator suboptimality), both SInd and DInd are worse but each contribute to the high performance of \ileed~in this multi-skill experiment.

\begin{table}[h!]
    \centering
    \caption{Performance of \ileed~Ablations}
    {\small \raggedright Mean of episodic reward over $20$ trials for the best demonstrator (BestD), BC, \ileed, as well as the state independent (SInd) and demonstrator independent (DInd) versions of \ileed.
    %\\Optimal (human): 0.25 \(\pm\) 0.11. Random: -12.30 \(\pm\) 0.11
    }
    \label{tab:ablations}
    \begin{tabular}{cccccc}
        \toprule
        $\beta$ & \textbf{BestD} & \textbf{BC} &  \textbf{SInd} & \textbf{DInd} & \textbf{ILEED}\\
        \midrule
        $.01$ & $0.44$ & $0.10$ & $0.19$ & $0.10$ & $\mathbf{0.70}$\\
        $.20$ & $0.76$ & $0.68$ & $0.74$ & $0.64$ & $\mathbf{0.86}$\\
        $1.0$ & $\mathbf{0.94}$ & $0.88$ & $0.89$ & $0.87$ & $0.89$\\
        \bottomrule
    \end{tabular}
\end{table}
\FloatBarrier

\section{Proofs} \label{sec:appendix_proofs}

\textbf{Proof of Proposition~\ref{prop:proper}}
\begin{proof}
% We let \(\phi\) and \(\bomega\) take on their ground truth values \(\phi^\star\) and \(\bomega^\star\), and expand out the inner expectation over the ground truth demonstrator policy:
We write the expectation as sampling first a demonstrator \(i\) and a state \(s\), and then sampling an action from the ground truth policy \( \pi(a | s, \omega^\star_i, \phi^\star, \pi_{\theta^\star}) \):
\begin{align*}
    \loss(\theta) &= - \max_{\phi, \bomega} \mathbb{E}_{i,s} \mathbb{E}_{\pi(a | s, \omega^\star_i, \phi^\star, \pi_{\theta^\star})} \Big[ \log \pi(a | s, \omega_i, \phi, \pi_\theta) \Big]
\end{align*}

The inner expectation corresponds to the log-loss scoring rule,
%for \( \omega^\star_i, \phi^\star, \pi_{\theta^\star}\)
which we know is strictly proper. Hence, \(\omega^\star_i, \phi^\star, \theta^\star\) maximizes the log term of each inner expectation. Therefore, by taking the \(\max\) over \(\phi, \omega\), we have that \(\theta^\star\) is a minimizer of \(\loss(\theta)\). (But may not be the unique minimizer since \(\phi\) and \(\bomega\) do not necessarily have to take on the values \(\phi^\star, \omega^\star\)).
\end{proof}

\textbf{Proof of Proposition~\ref{prop:bc}}
\begin{proof}
Since \(\loss(\theta)\) is proper (Proposition~\ref{prop:proper}) and \(\Theta\) is well-specified, we know that at \(\min_\theta \loss(\theta)\) we have:
\begin{align*}
    \min_\theta \loss(\theta) &= - \mathbb{E}_{i,s} \mathbb{E}_{\pi(a | s, \omega^\star_i, \phi^\star, \pi_{\theta^\star})} \Big[ \log \pi(a | s, \omega^\star_i, \phi^\star, \pi_{\theta^\star}) \Big]
\end{align*}

On the contrary, behavioral cloning uses a single policy to model all demonstrators.
\begin{align*}
    \loss(\theta_{BC}) &= - \mathbb{E}_{i,s} \mathbb{E}_{\pi(a | s, \omega^\star_i, \phi^\star, \pi_{\theta^\star})} \Big[ \log \pi_{\theta_{BC}}(a | s) \Big]
\end{align*}

Since the inner log-loss is a strictly proper loss function, the equality \(\min_\theta \loss(\theta) = \loss(\theta_{BC})\) only holds when \(\pi_{\theta_{BC}}(a | s) = \pi(a | s, \omega^\star_i, \phi^\star, \pi_{\theta^\star})\) for all \(i\), meaning that all the demonstrators must have identical policies.

\end{proof}

\textbf{Proof of Lemma~\ref{lem:strictlyproper}}
\begin{proof}

We need to show that under the conditions in Lemma~\ref{lem:strictlyproper},
\begin{align*}
    \loss_{\phi}(\theta^\star) &= - \max_{\omega} \mathbb{E}_{i, (s,a)} \Big[ \log \pi(a | s, \omega_i, \phi, \pi_{\theta^\star}) \Big] \\
    < \loss_{\phi}(\theta^\prime) &= - \max_{\omega} \mathbb{E}_{i, (s,a)} \Big[ \log \pi(a | s, \omega_i, \phi, \pi_{\theta^\prime}) \Big]
\end{align*}

for all \(\theta^\prime \neq \theta^\star\), where demonstrations \((s,a)\) for demonstrator \(i\) are drawn from \(\pi(a | s, \omega_i, \phi, \pi_{\theta^\star})\).

Again using the fact that log-loss is strictly proper, the equality \(\loss_{\phi}(\theta^\star) = \loss_{\phi}(\theta^\prime)\) only holds when the inner policies are equivalent for all states, i.e., \(\pi(a | s, \omega_i, \phi, \pi_{\theta^\star}) = \pi(a | s, \omega^\prime_i, \phi, \pi_{\theta^\prime})\) for all \(i, s, a\).

Recall that in the discrete action model:
\begin{align*}
\pi(a|s,\omega_i, \phi, \pi_{\theta^\star}) = \, \rho_\phi(s, \omega_i) \pi_{\theta^\star}(a | s)  + 
	 \frac{1-\rho_\phi(s, \omega_i)}{|\Actions|}
\end{align*}
and in the continuous action model:
\begin{align*}
    \pi(a | s, \omega_i, \phi, \pi_{\theta^\star}) = \sum_{j=1}^{k} \alpha_j \mathcal{N}(a ; \mu_j(s), \sigma_j(s) / \rho_\phi(s,\omega_i))
\end{align*}

To simplify notation, we will write the policy \(\pi(\cdot | \cdot, \omega_i, \phi, \pi)\) as \(\textsc{noise}(\pi, \rho_\phi(s,\omega_i))\).
Conveniently, from the form of either the discrete/continuous action model, we see that \(\textsc{noise}(\pi_{\theta^\star}, \rho_\phi(s,\omega_i)) = \textsc{noise}(\pi_{\theta^\prime}, \rho_\phi(s,\omega_i^\prime))\) if and only if \(\pi_{\theta^\prime} = \textsc{noise}(\pi_{\theta^\star}, \frac{\rho_\phi(s,\omega_i)}{\rho_\phi(s,\omega_i^\prime)})\), since the noise is multiplicative. 
In other words, an incorrect policy \(\pi_{\theta^\prime}\) can achieve optimal loss iff it is a noised version of the true policy \(\pi_{\theta^\star}\), and that the ratio of the expertise levels correspond to the noise of \(\pi_{\theta^\prime}\) relative to \(\pi_{\theta^\star}\). Moreover, this also tells us that the ratio must be the same for all demonstrators, so \(\frac{\rho_\phi(s,\omega_i)}{\rho_\phi(s,\omega_i^\prime)} = \frac{\rho_\phi(s,\omega_j)}{\rho_\phi(s,\omega_j^\prime)}\) for all \(i,j\).

However, when the expertise levels are intertwined, it may not be possible to set each expertise level to correspond to the desired noise. Recall that the expertise levels \(\rho_\phi(s,\omega) = \sigma(\langle f_\phi(s), \omega \rangle)\).

Suppose that at a state \(s\)
\begin{align*}
  f_\phi(s_0) = \alpha_1 f_\phi(s_1) + \alpha_2 f_\phi(s_2) + \ldots + \alpha_r f_\phi(s_r)
\end{align*}%

Then if \(\textsc{noise}(\pi_\theta, \rho_\phi(s,\omega)) = \textsc{noise}(\pi_{\theta^\prime}, \rho_\phi(s,\omega^\prime))\) for all states \(s_{0:m}\), we have that
\begin{align*}
\forall i, j, k : \frac{\rho_\phi(s_k,\omega_i)}{\rho_\phi(s_k,\omega_i^\prime)} &= \frac{\rho_\phi(s_k,\omega_j)}{\rho_\phi(s_k,\omega_j^\prime)}%\\
% \frac{\sigma(\langle s_k, \omega_i \rangle)}{\sigma(\langle s_k, \omega^\prime_i \rangle)} &= \frac{\sigma(\langle s_k, \omega_j \rangle)}{\sigma(\langle s_k, \omega^\prime_j \rangle)}\\
% \frac{\sigma(\langle s_k, \omega_i \rangle)}{\sigma(\langle s_k, \omega_i + C_i \rangle) } &= \frac{\sigma(\langle s_k, \omega_j \rangle)}{\sigma(\langle s_k, \omega_j + C_j \rangle) }\\
\end{align*}%

Or equivalently,
\begin{align*}
\forall i,  k &: \frac{\rho_\phi(s_k,\omega_i)}{\rho_\phi(s_k,\omega_i^\prime)} = C_k%\\
%\forall i,  k &: \frac{\sigma(\langle s_k, \omega_i \rangle)}{\sigma(\langle s_k, \omega^\prime_i \rangle) } = C_k
\end{align*}
Next we will rewrite to isolate the inner product.
\begin{align*}
\rho_\phi(s_k,\omega^\prime_i) &= \rho_\phi(s_k,\omega_i) / C_k \\
\sigma(\langle f_\phi(s_k), \omega^\prime_i \rangle) &= \rho_\phi(s_k,\omega_i) / C_k \\
\langle f_\phi(s_k), \omega^\prime_i \rangle &= \sigma^{-1} ( \rho_\phi(s_k,\omega_i) / C_k )
\end{align*}%
Now we can apply the linear dependence between \(s_0\) and \(s_{1:r}\) to get a relationship between the constants \(C_{0:r}\). In particular, for all \(i\):
\begin{align*}
\langle f_\phi(s_0), \omega^\prime_i \rangle &= \sum_{k=1}^{r} {\alpha_k \langle f_\phi(s_k), \omega^\prime_i \rangle}\\
\sigma^{-1} ( \rho_\phi(s_0,\omega_i) / C_0 ) &= \sum_{k=1}^{r} {\alpha_k  \sigma^{-1} ( \rho_\phi(s_k,\omega_i)) / C_k ) }\\
\end{align*}

Therefore, if no settings of the constants \(C_{0:r}\) can satisfy the above conditions, then we cannot set the expertise levels to accommodate a noised version of the true optimal policy \(\pi_{\theta^\star}\). In such a case, \(\theta^\star\) is the unique minimizer of \(\loss(\theta)\).

\end{proof}

\section{Implementation Details} \label{sec:appendix_implementation_details}
Overall we performed $3$ sets of experiments corresponding to the simulated MiniGrid environments, the continuous control task Square from the Robomimic study, and chess. We go over these three sets of experiments, detailing the framework and parameter setup used to run our method \ileed~as well as any other baselines we used for comparison. We note that working implementations of \ileed~and BC are provided in our supplementary material, with code that can be used to reproduce the MiniGrid results. For GAIL, we utilized an implementation built on top of rllab's codebase~\cite{duan2016benchmarking}, where the specific installation instruction are provided in the supplementary material. For experiments utilizing the Robomimic study we refer readers to the original codebase~\cite{robomimic2021}, noting that to implement \ileed~on top of their GMM policy class required minor changes to their framework which we detail below. \par
\begin{table}[!h]
    \small
    \caption{Parameters used for PPO}
    \raggedright Default implementation provided by stable baselines 3~\cite{stable-baselines3}, where below we list the specific parameters we changed.
    
	\label{tab:imp_ppo}
	\centering
	\begin{tabular}{cc}
		\toprule
		\multicolumn{1}{c}{Parameter} &
		\multicolumn{1}{c}{Description}\\
        \midrule
        Policy Class & MlpPolicy \\
        Update Steps & $128$\\
        Num. of Environments & $8$\\
        Batch Size & $4$\\
        Learning Rate & $0.00025$\\
        Timesteps & $200000$\\
        \bottomrule
	\end{tabular}
\end{table}

\textbf{MiniGrid Experiments}\\
For all MiniGrid experiments we relied on the provided Gym implementation~\cite{gym_minigrid} to simulate the environments, and used the well known stable baselines library~\cite{stable-baselines3} to train our policies with PPO. We list the specific hyperparameters in Table~\ref{tab:imp_ppo}, where we note that we used flattened observations for all MiniGrid experiments, relying on the standard `MlpPolicy' class provided by stable baselines. To run \ileed~and BC on the MiniGrid environments as discussed in Section~\ref{sec:results_mini} and~\ref{sec:results_multi} we utilized our own policy class consisting of a $3$--layer MLP with $4$ neurons in the hidden layer, ReLU non-linearities between, and a softmax at the output. For training, we ran $2000$ iterations utilizing two Adam~\cite{kingma2017adam} optimizers, one for parameters $\theta,\phi$, and $\psi$ with a learning rate of 1e-3, and the other for the expertise parameters $\omega$ with a smaller learning rate of 1e-2. For all experiments we relied on a two dimensional state-embedding $\fembed$ which was paremetrized by a $3$--layer MLP just like the policy. We list these parameters in Table~\ref{tab:imp_ileed} below, noting that our implementation is provided as part of the supplementary material. The parameters for GAIL are listed separately in Table~\ref{tab:imp_gail}, where we note that we utilized the recommended set of parameters provided by the implementation.

\begin{table}[!h]
    \small
    \caption{Parameters used for \ileed}
    \raggedright Implementation parameters for \ileed~are listed below. For BC we utilized the same parameters, removing the unnecessary components. 

	\label{tab:imp_ileed}
	\centering
	\begin{tabular}{cc}
		\toprule
		\multicolumn{1}{c}{Parameter} &
		\multicolumn{1}{c}{Description}\\
        \midrule
        Policy & $3$--layer MLP + Softmax \\
        Activation & ReLU \\
        Hidden Size(s) & $4$ \\
        Num. Iterations & $2000$\\
        Learning Rate $\theta,\phi,\psi$ & $0.001$\\
        Learning Rate $\omega$ & $0.01$\\
        State Embedding Dimension & $2$\\
        \bottomrule
	\end{tabular}
\end{table}

\begin{table}[!h]
    \small
    \caption{Parameters used for GAIL}
    \raggedright Implementation parameters for GAIL are listed below. 

	\label{tab:imp_gail}
	\centering
	\begin{tabular}{cc}
		\toprule
		\multicolumn{1}{c}{Parameter} &
		\multicolumn{1}{c}{Description}\\
        \midrule
        Policy & Categorical MLP\\
        Hidden Size(s) & $32,32$\\
        Latent Dimension & $2$\\
        Batch Size & $8000$\\
        Critic & Wassertstein \\
        Critic Epochs & $50$\\
        Critic Learning Rate & $0.0001$\\
        Critic Dropout Prob. & $0.6$\\
        Critic Penalty & $1$\\
        Critic Gradient Norm & $50$\\
        Recognition Epochs & $50$\\
        Recognition Learning Rate & $.0001$\\
        Scheduler $k$ & $20$\\
        TRPO Step Size & $0.01$\\
        \bottomrule
	\end{tabular}
\end{table}

\textbf{Robomimic Experiments}\\
As stated in the main text, we utilized the implementation provided by the original study~\cite{robomimic2021}, creating an instance of \ileed~by changing the RNN-GMM policy class to match the annotator model we defined in Eq.~\ref{eq:annotator_continuous}. Specifically, we edited the file \textit{robomimic/models/policy\_nets.py}, changing the \textit{RNNGMMActorNetwork} class by adding an optional scaling parameter to the variance which is outputted by the policy. We then utilized a separate Adam optimizer with a learning rate of 1e-4 to learn this parameter as done in the MiniGrid implementation. Although we do not include this specific implementation of \ileed~in our supplementary material due to the size of the Robomimic library, we plan to upload our algorithm directly to the Robomimic codebase. To run our experiments, we utilize the exact same setup as the original study. 

\textbf{Chess Experiments}\\
For the experiments on chess, we detail the architecture and parameters in Table~\ref{tab:imp_chess} below. \par

\begin{table}[!h]
    \small
    \caption{Parameters used for \ileed~on chess}
    \raggedright Implementation parameters for \ileed~when used with chess are listed below. For BC we utilized the same parameters, removing the unnecessary components. 

	\label{tab:imp_chess}
	\centering
	\begin{tabular}{cc}
		\toprule
		\multicolumn{1}{c}{Parameter} &
		\multicolumn{1}{c}{Description}\\
        \midrule
        Policy & $3$--layer MLP + Softmax \\
        Activation & ReLU \\
        Hidden Size(s) & $8$ \\
        Num. Iterations & $4000$\\
        Learning Rate $\theta,\phi,\psi$ & $0.001$\\
        Learning Rate $\omega$ & $0.01$\\
        State Embedding Dimension & $2$\\
        \bottomrule
	\end{tabular}
\end{table}

\FloatBarrier

\textbf{Computational Resources}\\
All of the experiments were performed on a machine with the 8C/16T Intel-9900K CPU, 32GB RAM, and an RTX-3080 GPU.

\begin{figure*}[!t]
    \centering
    \includegraphics[width=0.935\linewidth]{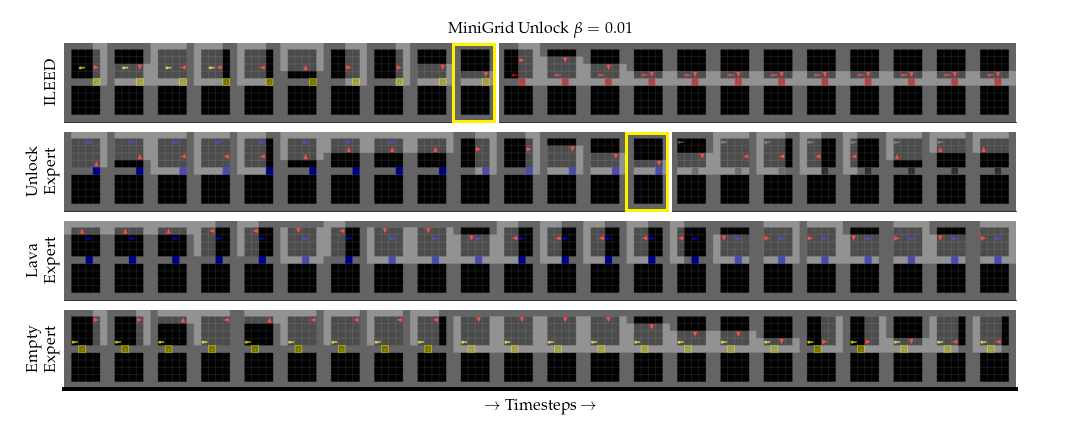}
    \includegraphics[width=0.935\linewidth]{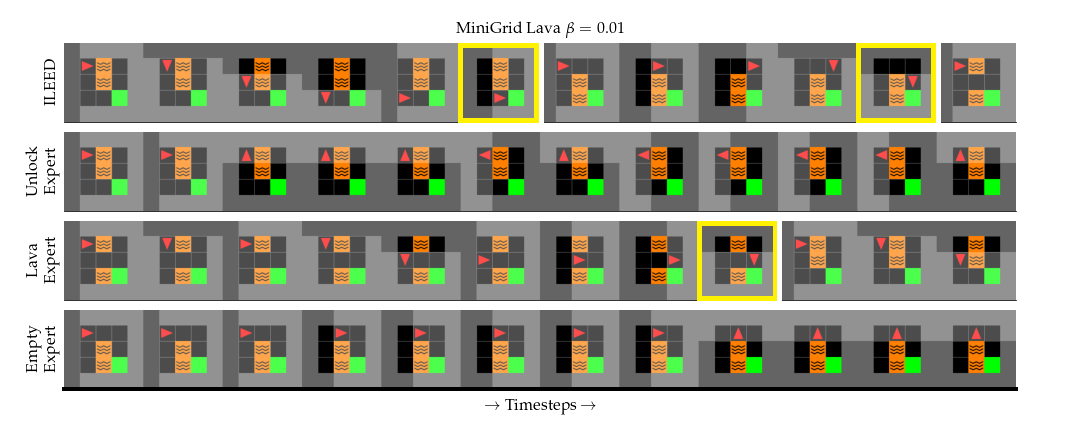}
    \includegraphics[width=0.935\linewidth]{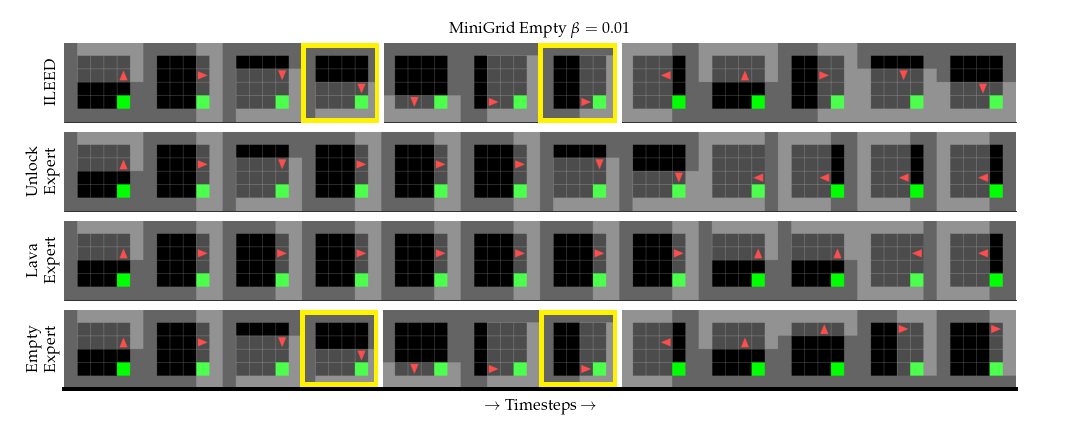}
    \caption{Trajectories corresponding to the MiniGrid experiments performed in Section~\ref{sec:results_multi}. For each environment, we show the trajectory of the policy learned by \ileed~on top and place the demonstrator trajectories below. The yellow bordered frames signify when the agent has successfully reached the goal. Since $\beta=0.01$ corresponds to the nosiest setting, we see the demonstrators only act optimally in the environment they are skilled in. For example, we see that other than \ileed, the \textit{Lava} expert is the only one to succeed in \textit{Lava}. On the other hand, we see \ileed~performing adequately in \textit{Lava} and \textit{Empty}, even showing some performance in the more challenging \textit{Unlock} environment despite the poor quality of the dataset. Specifically we see for the \textit{Unlock} environment, the policy trained by \ileed~is able to unlock the yellow door, but fails to unlock the red door. We emphasize again that the results shown are for the worst setting of $\beta=0.01$. As we increase $\beta$ the demonstration quality improves, making it easier for \ileed~to imitate multiple skills as shown in Table~\ref{tab:multi_minigrid}.}
    \label{fig:traj}
\end{figure*}
\end{document}